\documentclass[journal]{IEEEtran}

\usepackage{amsmath,amsthm,amssymb}
\usepackage{graphicx}                    
\usepackage[usenames,dvipsnames]{xcolor} 
\usepackage{comment}
\usepackage{subfigure} 
\usepackage{siunitx}                     
\usepackage[ruled,lined,linesnumbered]{algorithm2e}
\usepackage{cite}
\usepackage{array}
\usepackage{flushend}
\usepackage[hidelinks]{hyperref}

\newcommand{\tx}{\textnormal}

\newlength \figwidth
\newtheorem{Remark}{Remark}

\newtheorem{Lemma}{Lemma}

\newtheorem{Conjecture}{Conjecture}

\graphicspath{{./figures/}}

\newcolumntype{+}{>{\global\let\currentrowstyle\relax}}
\newcolumntype{^}{>{\currentrowstyle}}
\newcommand{\rowstyle}[1]{\gdef\currentrowstyle{#1}%
  #1\ignorespaces
}
\begin{document}

\title{Critical Zones for Comfortable Collision Avoidance with a Leading Vehicle}

\author{Jordanka Kovaceva, Nikolce Murgovski, Bal{\'{a}}zs Kulcs{\'{a}}r, Henk Wymeersch, Jonas B{\"{a}}rgman
\thanks{This research was supported by the Chalmers Area of Advance Transport project IRIS. \textit{(Corresponding author: Jordanka Kovaceva.)}}
\thanks{J. Kovaceva and  J. B{\"{a}}rgman are with the Department of Mechanics and Maritime Sciences, Chalmers University of Technology, SE-417 56 Gothenburg, Sweden (e-mail: jordanka.kovaceva@chalmers.se, jonas.bargman@chalmers.se).}
\thanks{N. Murgovski, B. Kulcs{\'{a}}r and H. Wymeersch are with the Department of Electrical Engineering, Chalmers University of Technology, Gothenburg, Sweden (e-mail: nikolce.murgovski@chalmers.se, kulcsar@chalmers.se, henkw@chalmers.se).}
}

\markboth{}%
{Kovaceva \MakeLowercase{\textit{et al.}}: Critical zones for comfortable collision avoidance with a leading vehicle}

\maketitle

\newpage

\begin{abstract}
This paper provides a general framework for efficiently obtaining the appropriate intervention time for collision avoidance systems to just avoid a rear-end crash. The proposed framework incorporates a driver comfort model and a vehicle model. We show that there is a relationship between driver steering manoeuvres based on acceleration and jerk, and steering angle and steering angle rate profiles. We investigate how four different vehicle models influence the time when steering needs to be initiated  to avoid a rear-end collision. The models assessed were: a dynamic bicycle model (DM), a steady-state cornering model (SSCM), a kinematic model (KM) and a point mass model (PMM). We show that all models can be described by a parameter-varying linear system. We provide three algorithms for steering that use a linear system to compute the intervention time efficiently for all four vehicle models. Two of the algorithms use backward reachability simulation and one uses forward simulation. 
Results show that the SSCM, KM and PMM do not accurately estimate the intervention time for a certain set of vehicle conditions. 
Due to  its fast computation time, DM with a backward reachability algorithm can be used for rapid offline safety benefit assessment, while DM with a forward simulation algorithm is better suited for online real-time usage.
\end{abstract}

\begin{IEEEkeywords}
Advanced driver assistance systems (ADAS), vehicle models, driver models, automotive safety.
\end{IEEEkeywords}

\section{Introduction}\label{sec:Intro}
\IEEEPARstart{A}{ccording} to the Annual Accident Report by the European Commission \cite{EC2021}, road traffic crashes cause 22700 fatalities and 1.2 million injuries each year in Europe. Similar numbers are reported for the United States by NHTSA \cite{NHTSA2021}. Advanced driver assistance systems (ADASs) are being developed and introduced to the market in order to reduce the number of crashes. ADAS helps the driver avoiding collisions with other road users such as vehicles, pedestrians, and cyclists. 
A specific type of ADAS, automated emergency braking and steering systems, detect an imminent collision and autonomously brake or perform an evasive steering manoeuvre to avoid the collision or mitigate its severity.  One of the challenges for these systems is deciding when to brake and when to steer. The ADAS needs to intervene early enough to avoid a crash, while acting too early may cause the driver  to consider the intervention a nuisance, as he or she still may be able to comfortably avoid a collision. Indeed, the safety benefit and acceptance of ADAS depend strongly on the time it issues an intervention \cite{Lubbe:2015}. 

The safety benefit of ADAS can be investigated retrospectively or prospectively. 
A retrospective assessment is based on real-world performance data, so the ADAS must be on the market long enough to be able to provide sufficient data for analysis   \cite{Cicchino:2017, Doyle2015, Isaksson-Hellman:2016, Kuehn:2009}. Alternatively,  a prospective assessment may be performed to evaluate new systems that are not yet on the market. This assessment is often done virtually (see, for example, \cite{Page:2015}) such as with counterfactual simulations \cite{Alvarez2017, Bargman:2017, Sander:2018}. The accuracy and computational complexity of these simulations depend on the driver, vehicle, and ADAS models, as well as on the details of the ADAS algorithm used to compute the  intervention time. 

Different algorithms use different strategies to initiate interventions from ADAS (for a comprehensive review see, for example, \cite{Dahl:2019}). Some consider time to collision (TTC) thresholds as the trigger, for example \cite{Chen2014}, others use the point of no return (triggering just before a crash is unavoidable) \cite{Hillenbrand2006}. Another option is to include driver comfort zone boundaries in the algorithm \cite{Sander:2017}. With this strategy, the ADAS does not intervene when the driver of the host (ego) vehicle could still comfortably avoid the crash; instead, triggering only when the comfort zone boundary has been crossed \cite{Sander:2017, Summala:2007}.  Then  ADAS evasive action can be initiated when the driver has passed the comfort zone boundary, for avoiding the collision by braking or steering for a specific set of initial vehicle conditions. The vehicle is then said to be in a so-called \textit{critical~zone} \cite{chandru17}. In particular, the critical zone boundary gives the latest ADAS intervention time for different initial conditions of the ego vehicle, e.g., the shortest TTC before collision avoidance becomes uncomfortable.

The authors in \cite{Brannstrom2014, Brannstrom2010} calculate the critical zone over a set of relative positions and relative speeds between the ego and leading vehicle, assuming that other ego vehicle states are initialised at zero. However, the critical zone depends also on other ego vehicle states such as yaw angle, yaw rate, steering angle, etc. 
Threat assessment algorithms that calculate the critical zone for all of the above mentioned initial conditions have not yet been investigated in literature.

With respect to the definition of comfort, several studies quantify the levels of acceleration and jerk that drivers consider comfortable, both for braking and steering \cite{Bargman:2015, Kiefer:2003}. The vehicle control actions performed by the driver, taken together, define an action profile, which may take the form of piece-wise constant acceleration and jerk for braking and steering \cite{Fambro2000, LeBlanc2017} or piece-wise constant steering angle or steering angle rate \cite{godthelp86}. The former will hereafter be referred to as the \textit{acceleration-jerk (AJ) profile}, while the latter will be referred to as the \textit{steering angle-steering rate (SASR) profile}. For both profiles, the values of the acceleration, jerk, steering angle, and steering angle rate can be based on vehicle performance boundaries (how hard the vehicle can brake or steer) \cite{Kuehn:2009}, driver comfort boundaries \cite{Sander:2018}, or limits set by choosing values from very rare occurrences of the variables in real driving \cite{Brannstrom2010}.

In addition to the driver models, the time the ADAS issues an intervention also depends  on the vehicle model. To reduce false interventions, the simulations should be performed with accurate, and possibly complex, vehicle models to be as similar as possible to the real world. However, the computation time of these simulations may increase rapidly with the model complexity \cite{Kalra:2016, Dahl:2019}. One way to keep computations tractable and  enable rapid safety benefit assessment is to use simplified vehicle models and efficient computational  methods (for computing the intervention time, for example).

The simplest vehicle model is arguably the point mass vehicle model (PMM), which has been used for generating ADAS steering profiles \cite{Shiller:1998}, \cite{NilssonJ:15}. The advantage of using this model is that because of its linearity, it is straightforward to obtain a closed-form analytical solution for the intervention time. This significantly reduces computational complexity, but it also reduces accuracy and thus carries a risk of too early or too late interventions. Furthermore, PMMs do not include a steering angle signal and thus can only be used with AJ profiles \cite{chandru17}, making it impossible to investigate the influence of SASR profiles on, for example, the time of ADAS intervention.

In contrast, SASR profiles can be used with more detailed vehicle models, such as the single-track linear steady state cornering model (SSCM), also known as the bicycle SSCM \cite{Gillespie1992}. This model has been investigated in, for example, \cite{Brannstrom2010}; although a closed-form analytical solution for computing the intervention time was not provided. To the best of the authors' knowledge, the most advanced vehicle model for computing critical zones in commercial ADAS is the SSCM.

Other single-track linear vehicle models that can be used with the SASR profiles are the kinematic model (KM) and the dynamic model (DM) \cite{rajamani06}. The KM has an accuracy in between those of PMM and SSCM, while the DM is more complex than the SSCM~\cite{rajamani06}, which may increase the accuracy of its calculations for the  intervention time. However, the KM and DM have not been used before to calculate the time of ADAS intervention, as far as we can ascertain,  nor is it clear what their computational complexity is for computing the intervention time.

Most previous studies have used just one type of driver model, either AJ or SASR, with one type of vehicle model, typically PMM or SSCM. This makes it difficult to understand how the same driver model is realised with different vehicle models, and how different vehicle models compare when using the same driver model. 

There have been efforts to benchmark different vehicle models in vehicles motion planning  \cite{Althoff:2017}, but as far as the authors can ascertain, there is a research gap due to a lack of studies either investigating what type of effect the different vehicle models have on the ADAS intervention time or comparing the different models' benefits in counterfactual simulations.

This paper aims to close these knowledge gaps and provide methods for efficiently computing intervention time for a wide set of initial ego vehicle conditions, including position, speed, yaw angle, yaw rate and steering angle. This paper limits its scope to only particular ADAS components related to the computation of intervention time and does not aim to develop the entire ADAS.  
Specifically, this paper contributes to the research field by: 
\begin{itemize}
    \item proposing a general framework for computing intervention time for evasive steering manoeuvres using different initial conditions and different complexity levels of vehicle modelling;
    \item deriving the relationship between AJ and SASR profiles in rear-end collision scenarios;
    \item developing computationally efficient algorithms for obtaining the intervention time, using the KM, SSCM, and DM vehicle models;
    \item benchmarking the computational complexity and accuracy for computing intervention time using single-track linear vehicle models based on PMM, KM, SSCM, and DM.
\end{itemize}

This paper is organised as follows. Section~\ref{sec:ModellingAndProblem} describes the problem and the modelling approach. Section~\ref{sec:avoidance_braking} describes an algorithm for avoiding collision by braking, while Section~\ref{sec:avoidance_steering} describes an algorithm for avoiding collision by steering. Section~\ref{sec:benchmarkModels} describes the simplified vehicle models used for the benchmark and Section~\ref{sec:Results} shows the results. Section~\ref{sec:Discussion} discusses the differences among the models and the limitations of the study as well as providing an outlook on future research. Finally, Section~\ref{sec:Conclusion} presents the conclusions.

\section{Modelling and problem statement} \label{sec:ModellingAndProblem}
This section formulates the problem of obtaining the intervention time, before which the driver in the ego vehicle would need to act (by braking or steering) to comfortably avoid a collision with a leading vehicle. The intervention time is the point in time when the ADAS threat assessment algorithm decides the system should intervene: for the comfort zone boundary approach, the system's evasive action is initiated when the driver has passed a comfort boundary. 
This section also describes mathematical models of the leading and ego vehicles in a rear-end conflict situation and the typical control model of the ego vehicle driver. In this paper, the longitudinal and lateral dynamics of the ego vehicle are decoupled, as is typical in the computation of ADAS intervention time \cite{Brannstrom2010, Brannstrom2014, chandru17, Eidehall:2011, Eidehall:2013}. 

\subsection{Model of the leading vehicle}

\begin{figure}[t]
	\centering
        \includegraphics{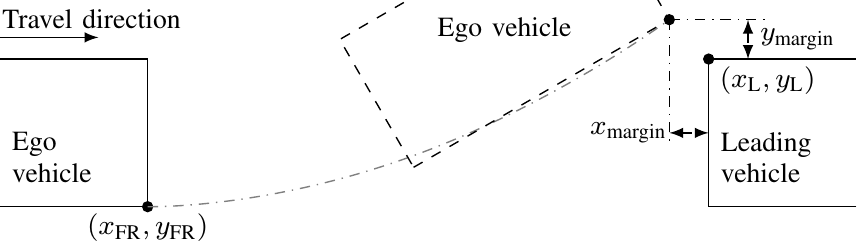}
	\caption{Illustration of an ego vehicle avoiding collision with a leading vehicle by steering. Collision is avoided if the front right corner of the ego vehicle is above the rear left corner of the leading vehicle (shifted by given safety margins).}
	\label{fig:collision_avoidance}
\end{figure}

The collision avoidance manoeuvre of the ego vehicle is described under certain assumptions about the future motion of the leading vehicle. The most common assumption is that the leading vehicle will continue driving with the same longitudinal speed $v_\tx{L}$ and that it will maintain the same lateral position \cite{Shamir2004}. Let $x_\tx{L}(0)$ denote the initial longitudinal position of the leading vehicle rear end; see Fig.~\ref{fig:collision_avoidance}. 
Then, at any point in time $t$, its longitudinal position progresses according to
\begin{align}
    x_\tx{L}(t) = x_\tx{L}(0) + v_\tx{L} t.
\end{align}
The lateral position of the leading vehicle's rear left corner, $y_\tx{L}$, remains constant (i.e., $y_\tx{L}(0)=y_\tx{L}(t)$).

This general assumption is used when assessing safe overtaking on highways and rural roads; see, for example, \cite{Shamir2004}. Later, in Section~\ref{sec:Discussion} it is discussed how the proposed algorithm can be applied to scenarios where the leading vehicle lateral speed is not zero and when its longitudinal speed is uncertain.

\subsection{Braking dynamics of ego vehicle} \label{sec:long-dynamics}
\begin{figure}[t]
	\centering
        \includegraphics{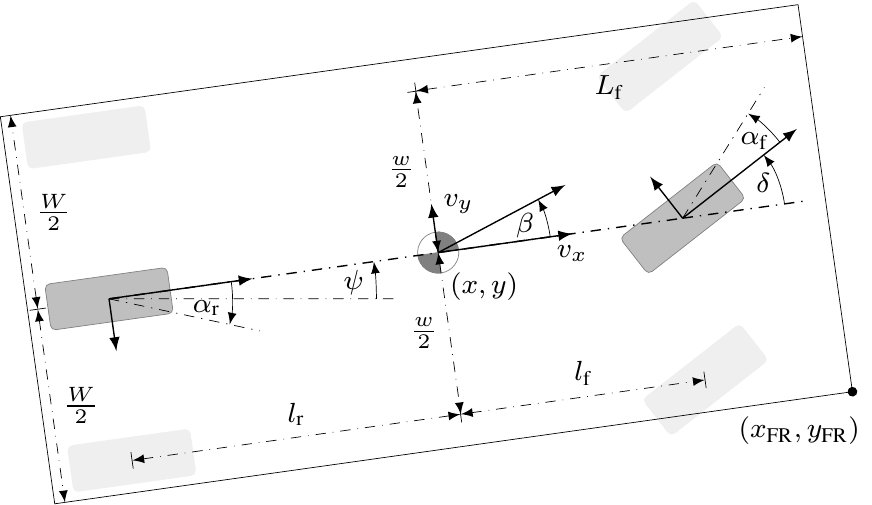}
	\caption{Front-wheel steering vehicle model. The model is approximated with a single track by grouping the rear wheels together in  the middle of the rear axle, and the front wheels in the middle of the front axle.}
	\label{fig:dynamic-bicycle-model}
\end{figure}

In threat assessment studies with  decoupled longitudinal and lateral vehicle dynamics, the ego vehicle's longitudinal motion is generally described by a point mass system to model braking action only \cite{Brannstrom2010}. To avoid a collision with the leading vehicle, the ego vehicle's frontmost point must be behind the rearmost point of the leading vehicle by at least a safety margin $x_\tx{margin}$. With the longitudinal position of the ego vehicle reference point denoted as $x_\tx{b}$ , the safety constraint can be described as
\begin{align} \label{eq:longitudinal-safety}
    x_\tx{b}(t) + L_\tx{f} \leq x_\tx{L}(t) - x_\tx{margin}
\end{align}
where $L_\tx{f}$ is distance from the reference point to the ego vehicle's front, see Fig.~\ref{fig:dynamic-bicycle-model}. 
Let 
\begin{align} \label{eq:delta_x}
    \Delta x_\tx{b}(t) = x_\tx{b}(t) - x_\tx{L}(t) + L_\tx{f}
\end{align}
denote the distance between the vehicles. Differentiating \eqref{eq:delta_x} with respect to time results in a similar relation between ego and leading vehicle speed. At the boundary condition, when the ego vehicle front is positioned just at the safety margin, it is required that for a certain braking time $t_\tx{b}$ the relative speed $\Delta v_\tx{b}(t)$ is reduced to zero,
\begin{align} \label{eq:long-safety-margin}
    &\Delta v_\tx{b}(t_\tx{b}) = v_\tx{b}(t_\tx{b}) - v_\tx{L} = 0.
\end{align}
Here, $v_\tx{b}$ is the longitudinal speed of the ego vehicle reference point; for the studied scenario, it must initially be greater than the leading vehicle speed: i.e., $v_\tx{b}(0)> v_\tx{L} > 0$. It is also required that $\Delta x_\tx{b}(0) \leq -x_\tx{margin}$. This algorithm, proposed later in Section~\ref{sec:avoidance_braking}, will also work when the leading vehicle stops abruptly (i.e., $v_\tx{L}=0$, then $v_\tx{b}(t_b)=0$).

\begin{figure}[t]
    \centering
    \includegraphics{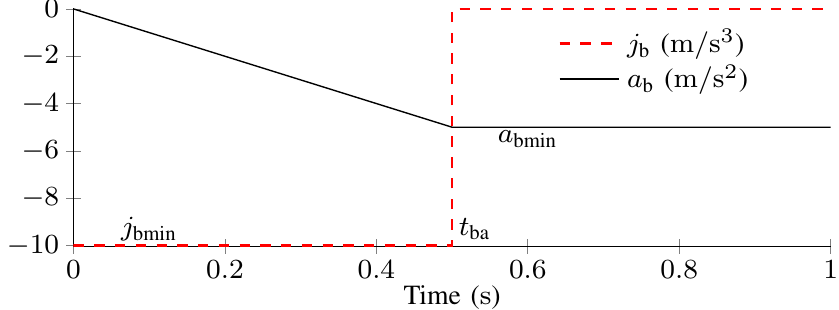}
\caption{Braking control model. The ego vehicle follows a constant longitudinal jerk $j_\tx{bmin}$ until time $t_\tx{ba}$ when it reaches the minimum longitudinal acceleration $a_\tx{bmin}$. The ego vehicle follows a constant acceleration from time $t_\tx{ba}$ onward.}
\label{fig:braking-profile}
\end{figure}

Informed by typical driver behaviour when avoiding a rear-end collision by braking, longitudinal jerk $j_\tx{b}$ is taken as the control input: $u_\tx{b}=j_\tx{b}$ \cite{Brannstrom2010, Brannstrom2014}. This widely used control model assumes that the driver maintains a minimum longitudinal jerk $j_\tx{bmin}<0$ until either the safety margin \eqref{eq:long-safety-margin} is reached or the minimum allowed longitudinal acceleration $a_\tx{bmin}<0$ is reached \cite{Fambro2000, Lee2006, McGehee1999, LeBlanc2017}. In the latter case, the driver continues braking with $a_\tx{bmin}$ until the safety margin \eqref{eq:long-safety-margin} is reached, see Fig.~\ref{fig:braking-profile}. 

To model the longitudinal motion, we define the state vector as 
\begin{align}
    \mathbf{x}_\tx{b} = \begin{bmatrix} \Delta x_\tx{b} & \Delta v_\tx{b} & a_\tx{b} \end{bmatrix}^\top
\end{align}
where $a_\tx{b}$ is the longitudinal acceleration of the ego vehicle reference point.
The braking dynamics can then be summarised by the linear system
\begin{align} \label{eq:dynamic-bicycle-longitudinal}
\dot{\mathbf{x}}_\tx{b}(t) = A_\tx{b} \mathbf{x}_\tx{b}(t) + B_\tx{b} u_\tx{b}(t)    
\end{align}
with the matrices defined as
\begin{align}
    A_\tx{b} = \begin{bmatrix} 0 & 1 & 0\\0 & 0 & 1\\0 & 0 & 0 \end{bmatrix}, \quad B_\tx{b} = \begin{bmatrix} 0\\0\\1 \end{bmatrix}.
\end{align}

\subsection{Ego vehicle steering dynamics} \label{sec:lat-danymics}
To describe lateral motion, the ego vehicle is modelled as a front-wheel-steered, single-track (bicycle) model; see Fig.~\ref{fig:dynamic-bicycle-model}. The vehicle course angle is given by $\beta+\psi$, where $\psi$ is the vehicle heading (yaw) angle and $\beta$ is the vehicle slip angle. The steering angle of the front wheel is denoted by $\delta$ and the lateral slip angles of the front and rear tires are given by $\alpha_\tx{f}$ and $\alpha_\tx{r}$, respectively. Its lateral dynamics, as proposed in \cite[pp.~27-39]{rajamani06}, are modelled under the following considerations:
\begin{itemize}
	\item The longitudinal vehicle velocity $v_x=v_\tx{b}(0)$ is constant;
	\item The vehicle is operated in the linear tire region without longitudinal slip;
	\item The wheels' moments of inertia and longitudinal tire forces can be neglected;
	\item Small angle approximations $\sin\delta \approx \delta$, ${\cos\delta\approx 1}$ are applied for all angles, including $\psi$, $\beta$, $\alpha_\tx{r}$ and $\alpha_\tx{f}$.
\end{itemize}
The ego vehicle lateral dynamics can then be described as 
\begin{subequations} \label{eq:dynamic-bicycle}
	\begin{align}
	    \dot y &= v_x\sin\psi + v_\tx{s}\cos\psi\approx v_x\psi + v_\tx{s}\\
		\dot v_\tx{s} &= - 2\frac{c_{\tx{f}} +c_{\tx{r}}}{m v_x} v_\tx{s} -  \left(v_x + 2 \frac{l_\tx{f}c_{\tx{f}}-l_\tx{r}c_{\tx{r}}}{m v_x}\right) \dot \psi + \frac{2 c_{\tx{f}}}{m}\delta\\
		\ddot \psi&= \frac{2}{I_z}\left(-\frac{l_\tx{f} c_{\tx{f}}-l_\tx{r} c_{\tx{r}}}{v_x} v_\tx{s} - \frac{l_\tx{f}^2 c_{\tx{f}} + l_\tx{r}^2 c_{\tx{r}}}{v_x} \dot \psi + l_\tx{f} c_{\tx{f}}\delta\right)
	\end{align}
\end{subequations}%
where $y$ is the lateral position of the vehicle's reference point in the inertial frame; $v_\tx{s}$ is the lateral speed of the reference point in the vehicle frame; $c_\tx{f}$ and $c_\tx{r}$ are lateral cornering stiffness coefficients of the front and rear tires, respectively; $l_\tx{f}$ and $l_\tx{r}$ are the distances from the reference point to the front and rear wheel axles, respectively; $m$ is the vehicle mass; and $I_z$ is the rotational moment of inertia about the vertical axis at the reference point \cite[pp.~27-39]{rajamani06}.

To avoid a collision with the leading vehicle (see Fig.~\ref{fig:collision_avoidance}), the front right vehicle corner must be beyond the safety margin of the leading vehicle. The lateral position of the front right corner can be expressed as 
	\begin{align} \label{eq:FR-coordinates}
		y_\tx{FR} &= y + L_\tx{f}\sin\psi - \frac{W}{2}\cos\psi \approx y + L_\tx{f}\psi - \frac{W}{2}
	\end{align}
where $W$ is the vehicle width. The safety constraint can then be described as
\begin{align}\label{eq:lat-safety-margin}
    y_\tx{FR}(t_\tx{s}) \geq y_\tx{L} + y_\tx{margin}
\end{align}
where $t_\tx{s}$ is the steering time and $y_\tx{margin}$ is a safety margin. 

To model evasive manoeuvres, previous studies have proposed two different steering control strategies, based on either the AJ or the SASR driver profile. Control models based on the AJ profile consider lateral jerk, generally assumed to be a piece-wise constant function \cite{chandru17, LeBlanc2017},  as the control input. Control models based on the SASR profile consider a piece-wise constant steering angle rate $\omega=\dot\delta$ as the control input \cite{Scanlon:2015}. Here, we consider the latter model; the control input is $u_\tx{s}=\omega$. (Later, in Section~\ref{sec:avoidance_steering}, we show this model's relation to the  AJ driver profile.) 

\begin{figure}[t]
    \centering
    \includegraphics{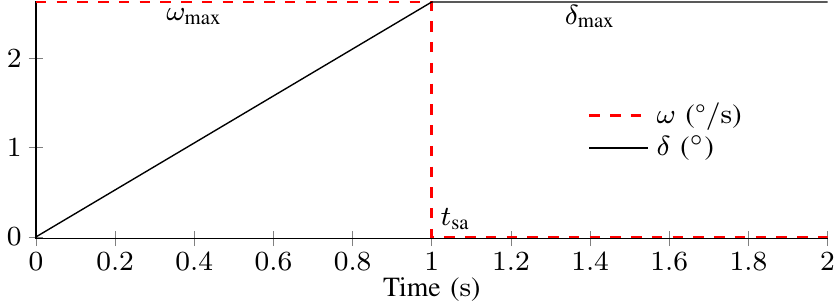}
\caption{Steering control model. The ego vehicle follows a constant steering angle rate $\omega_\tx{max}$ until time $t_\tx{sa}$ when it reaches maximum steering angle $\delta_\tx{max}$. The ego vehicle follows a constant steering angle from time $t_\tx{sa}$ onward.}
\label{fig:steering-profile}
\end{figure}

This model assumes that the shortest distance between the vehicles that still allows the driver to avoid a frontal collision by steering is achieved when the driver maintains a maximum steering rate $\omega_\tx{max}$ until either the safety margin \eqref{eq:lat-safety-margin} or the maximum allowed steering  angle $\delta_{\tx{max}}$ is reached \cite{Brannstrom2010, Brannstrom2014}. If the maximum steering angle $\delta_{\tx{max}}$ is reached first, then the driver maintains this angle until the safety margin \eqref{eq:lat-safety-margin} is reached; see Fig.~\ref{fig:steering-profile}. 
This manoeuvre is often referred to as a J-manoeuvre \cite{Brannstrom2010, LeBlanc2017}.  (Section~\ref{sec:Discussion} will discuss how the method can be extended to more complex manoeuvres such as S-manoeuvres, in which the ego vehicle is displaced laterally, but the vehicle heading is the same at the start and end of the manoeuvre \cite{He:2019, karlsson19}).

For the J-manoeuvre, let the lateral vehicle state and output vectors be
\begin{subequations}  \label{eq:vectors_bicycle_model}
\begin{align}
	\mathbf{x}_\tx{s}&=\begin{bmatrix} y & \psi & v_\tx{s} & \dot \psi & \delta \end{bmatrix}^\top\\
	\mathbf y_\tx{s} &= \begin{bmatrix} y_\tx{FR} + \frac{W}{2} & a_\tx{s} & j_\tx{s} \end{bmatrix}^\top,
\end{align}
\end{subequations} 
respectively, where 
\begin{subequations} 
\begin{align}
    a_\tx{s} &= \dot v_\tx{s} + v_x \dot\psi\\
    j_\tx{s} &= \ddot v_\tx{s} + v_x \ddot\psi
\end{align}
\end{subequations}
are lateral acceleration and jerk, respectively, at the reference point in the vehicle frame. Then, the lateral vehicle dynamics can be described by the parameter-varying linear system
\begin{subequations} \label{eq:dynamic-bicycle-lateral}
\begin{align}
	\dot{\mathbf x}_\tx{s}(t) &= A_\tx{s}(v_x)\mathbf x_\tx{s}(t) + B_\tx{s} u_\tx{s}(t)\\
	\mathbf y_\tx{s}(t) &= C_\tx{s}(v_x)\mathbf x_\tx{s}(t) + D_\tx{s}(v_x) u_\tx{s}(t)
\end{align}
\end{subequations} 
with the matrices $A_\tx{s}$, $B_\tx{s}$, $C_\tx{s}$, and $D_\tx{s}$ chosen appropriately. For a detailed description see Table~\ref{TT:ModelMatrices} in Appendix~\ref{sec:different-models}. 

\subsection{When can the driver avoid a collision comfortably?} \label{sec:problem_statement}
Let the vectors $\mathbf x_\tx{b}(0)$ and $\mathbf x_\tx{s}(0)$ denote the current ego vehicle state. The collision is safely avoided if the ego vehicle driver is able to either brake or steer away by engaging either of the control models discussed in Sections~\ref{sec:long-dynamics} and \ref{sec:lat-danymics}. In particular, let the function 
\begin{align} \label{eq:braking-time}
    t_\tx{b} = f_\tx{b}(\mathbf x_\tx{b}(0), u_\tx{b}(t))
\end{align}
provide the braking intervention time needed for the driver to reduce the relative speed $\Delta v_\tx{b}$ to zero. The relative longitudinal distance between the vehicles, $\Delta x_\tx{b}(t_\tx{b})$, is then computed by integrating (simulating forward) the system \eqref{eq:dynamic-bicycle-longitudinal} from time $0$ to $t_\tx{b}$. If ${-\Delta x_\tx{b}(t_\tx{b})\geq x_\tx{margin}}$, then collision is safely avoided. 

If the previous condition is not satisfied, it remains to be checked whether the ego vehicle driver is able to steer away to avoid a collision with the leading vehicle. Let the function 
\begin{align} \label{eq:steering-time}
    t_\tx{s} = f_\tx{s}(\mathbf x_\tx{s}(0), u_\tx{s}(t), v_x, y_\tx{L}, \mathbf p)
\end{align}
provide the steering intervention time needed   for the ego vehicle front right corner to reach 
	$y_\tx{FR}(t_\tx{s}) = y_\tx{L} + y_\tx{margin}$ 
 (given the steering control model in Fig.~\ref{fig:steering-profile} and values for $\delta_\tx{max}$ and $\omega_\tx{max}$ based either on driver comfort or vehicle performance).
Here $\mathbf p$ gathers the vehicle parameters that are constant during the manoeuvre; see Table~\ref{TT:ModelMatrices} in Appendix~\ref{sec:different-models}. 
The longitudinal distance travelled during the steering manoeuvre can be computed by integrating (simulating forward) the nonlinear system
\begin{align} \label{eq:dot_xs}
    \dot x_\tx{s} = v_x\cos\psi - v_\tx{s}\sin\psi \approx v_x - v_\tx{s}\psi
\end{align}
from time $0$ to $t_\tx{s}$, with $x_\tx{s}(0) = x_\tx{b}(0)$. The longitudinal position of the front right corner at time $t_\tx{s}$ can be expressed as $x_\tx{FR}(t_\tx{s}) = x_\tx{s}(t_\tx{s})  + L_\tx{f}\cos\psi(t_\tx{s}) + \frac{W}{2}\sin\psi(t_\tx{s})$. At time $t_\tx{s}$, the longitudinal distance between the leading vehicle and the ego vehicle front right corner is
\begin{align} 
    \begin{split} \label{eq:delta_xs}
        \Delta x_\tx{s}(t_\tx{s}) 
        &\approx x_\tx{s}(t_\tx{s}) + L_\tx{f} + \frac{W}{2}\psi(t_\tx{s}) - x_\tx{L}(t_\tx{s})
    \end{split}
\end{align}
where small angle approximation has been applied. If ${-\Delta x_\tx{s}(t_\tx{s})\geq x_\tx{margin}}$, then the collision is safely avoided. 

Clearly, if the functions $f_\tx{b}$ and $f_\tx{s}$ are known, the remaining steps for investigating the appropriate intervention time for comfortable collision avoidance are straightforward. Indeed, one of the goals of this paper is to obtain the functions $f_\tx{b}$ and $f_\tx{s}$ that provide the braking and steering intervention time to avoid a crash, given the control models in Figs.~\ref{fig:braking-profile}-\ref{fig:steering-profile} and a set of thresholds based either on driver comfort or vehicle performance. Moreover, to keep computations tractable and enable rapid safety benefit assessment, $f_\tx{b}$ and $f_\tx{s}$ should be computed efficiently (covered in Sections~\ref{sec:avoidance_braking} and \ref{sec:avoidance_steering}).

\section{Collision avoidance by braking} \label{sec:avoidance_braking}
This section describes the computation of the braking time $t_\tx{b}$ (i.e., the function $f_\tx{b}$) and the algorithm for avoiding a collision by braking.

Linear ordinary differential equations, as in \eqref{eq:dynamic-bicycle-longitudinal}, have a solution of the form  
\begin{align} \label{eq:ZOH}
    \mathbf x_\tx{b}(t) = e^{A_\tx{b} t} \mathbf x_\tx{b}(0) + \int_0^{t} e^{A_\tx{b}(t-\tau)} B_\tx{b} u_\tx{b}(\tau)\tx{d}\tau.
\end{align}
Since the longitudinal jerk is a piece-wise constant function, the integration can be performed in parts. For a constant longitudinal jerk $\bar u_\tx{b}$, the solution simplifies to 
\begin{align} \label{eq:brake-dynamics-discrete}
    \mathbf x_\tx{b}(t) = A_{\tx{b}t}(t) \mathbf x_\tx{b}(0) + B_{\tx{b}t}(t) \bar u_\tx{b}
\end{align}
with the matrices defined as
\begin{align}
    A_{\tx{b}t}(t) = \begin{bmatrix}1 & t & \frac{t^2}{2}\\
                            0 & 1 & t\\
                            0 & 0 & 1\end{bmatrix}, \quad 
    B_{\tx{b}t}(t) = \begin{bmatrix} \frac{t^3}{6}\\ \frac{t^2}{2} \\ t \end{bmatrix}.
\end{align} 

Consider the case when the constant control input is ${\bar u_\tx{b}=j_\tx{bmin}}$. 
\begin{Lemma}
    The time until the relative speed in \eqref{eq:brake-dynamics-discrete} drops to zero, with $\bar u_\tx{b}=j_\tx{bmin}$, can be obtained as 
    \begin{align} \label{eq:brake-time-jerk}
        t_\tx{bj} = \frac{-a_\tx{b}(0) - \sqrt{a_\tx{b}^2(0) - 2 j_\tx{bmin} \Delta v_\tx{b}(0)}}{j_\tx{bmin}}.
    \end{align}
\end{Lemma}

\begin{proof}
It follows directly from the second row in \eqref{eq:brake-dynamics-discrete} that obtaining the time for decreasing $\Delta v_\tx{b}(0)$ to zero requires solving a quadratic equation, of which \eqref{eq:brake-time-jerk} is one of the roots. Moreover, the discriminant in \eqref{eq:brake-time-jerk} is strictly greater than $a_\tx{b}^2(0)$, since ${j_\tx{bmin}<0}$ and $\Delta v_\tx{b}(0)>0$. This fact immediately shows that the other root of the quadratic equation gives a negative time, and hence is not feasible. 
\end{proof}

It remains to be checked whether the minimum acceleration is reached before the time $t_\tx{bj}$. The time to reach minimum acceleration can be computed as
\begin{align} \label{eq:brake-time-acceleration}
    t_\tx{ba} = \frac{a_\tx{bmin}-a_\tx{b}(0)}{j_\tx{bmin}}.
\end{align}
If $t_\tx{ba}\geq t_\tx{bj}$, then the braking time is obtained as $t_\tx{b}=t_\tx{bj}$. Otherwise, the system \eqref{eq:brake-dynamics-discrete} needs first to be solved until time $t_\tx{ba}$ with $\bar u_\tx{b}=j_\tx{bmin}$, to obtain the state vector $\mathbf x_\tx{b}(t_\tx{ba})$. Notice that the acceleration state at $t_\tx{ba}$ is equal to $a_\tx{bmin}$, and it is sufficient to solve only for the first two rows in \eqref{eq:brake-dynamics-discrete}. Taking $\mathbf x_\tx{b}(t_\tx{ba})$ as the initial state, system \eqref{eq:brake-dynamics-discrete} needs to be solved one more time until $t_\tx{b}$ with $\bar u_\tx{b}=0$. Yet again, the relative speed $\Delta v_\tx{b}(t_\tx{b})$ has to be zero, from which the braking time can be obtained as 
\begin{align}
    t_\tx{b} = t_\tx{ba} - \frac{\Delta v_\tx{b}(t_\tx{ba})}{a_\tx{bmin}}.     
\end{align}

Finally, the algorithm for avoiding a collision by braking can be summarised as in Algorithm~\ref{alg:fx}. The algorithm includes the closed-form solution of the function $f_\tx{b}$ sought in \eqref{eq:braking-time}, which computes the braking time $t_\tx{b}$.
\begin{algorithm}[tb]
\SetAlgoLined
\DontPrintSemicolon
  \textbf{Inputs}: $\mathbf x_\tx{b}(0)$, $x_\tx{L}(0)$, $v_\tx{L}$, parameters\;
  Compute $t_\tx{bj}$ and $t_\tx{ba}$ from \eqref{eq:brake-time-jerk} and \eqref{eq:brake-time-acceleration}, respectively\;
  \eIf{$t_\tx{ba}\geq t_\tx{bj}$}{
    $t_\tx{b}=t_\tx{bj}$\;
    $\mathbf x_\tx{b}(t_\tx{b}) = A_{\tx{b}t}(t_\tx{b}) \mathbf x_\tx{b}(0) + B_{\tx{b}t}(t_\tx{b}) j_\tx{bmin}$\;
    }  
    {$\mathbf x_\tx{b}(t_\tx{ba}) = A_{\tx{b}t}(t_\tx{ba}) \mathbf x_\tx{b}(0) + B_{\tx{b}t}(t_\tx{ba}) j_\tx{bmin}$\;
    $t_\tx{b} = t_\tx{ba} - \Delta v_\tx{b}(t_\tx{ba})/a_\tx{bmin}$\;
    $\mathbf x_\tx{b}(t_\tx{b}) = A_{\tx{b}t}(t_\tx{b}-t_\tx{ba}) \mathbf x_\tx{b}(t_\tx{ba})$\;
    }
    \eIf{$-\Delta x_\tx{b}(t_\tx{b})\geq x_\tx{margin}$}{
    Collision can be avoided by braking\;
    }
    {Collision cannot be avoided by braking\;
    }
 \caption{Collision avoidance by braking.}
 \label{alg:fx}
\end{algorithm}

\section{Collision avoidance by steering} \label{sec:avoidance_steering}
This section describes the relationship between the two steering strategies AJ and SASR, and shows how road friction influences steering performance. Then, using a control model based on the SASR profile, an algorithm is developed to test whether the driver may avoid a collision by steering. 

\subsection{Steering control model}
The relationship between the AJ and SASR strategies is investigated in steady-state cornering conditions of the dynamic bicycle model \eqref{eq:dynamic-bicycle}. Steady-state cornering conditions are reached when $\ddot\psi=0$ and $\dot\beta=0$, implying $\dot v_\tx{s}=0$ and $\dot\delta=0$; see, e.g., \cite[pp.~195-230]{Gillespie1992}. 
\begin{Lemma}
A steady-state cornering lateral acceleration $a_\tx{ss}$, experienced at the vehicle reference point, can be achieved by a constant steering angle
\begin{align} \label{eq:steady-state-angle}
\delta_\tx{ss}(a_\tx{ss},v_x) = \frac{a_\tx{ss}}{l}\left(\left(\frac{l}{v_x}\right)^2 + \frac{m}{2}\left(\frac{l_\tx{r}}{c_\tx{f}} - \frac{l_\tx{f}}{c_\tx{r}} \right)\right)
\end{align}
where $l=l_\tx{r}+l_\tx{f}$ denotes the wheel base. Thus, a steady lateral jerk $j_\tx{ss}$ at the vehicle reference point can be achieved by a constant steering rate 
\begin{align} \label{eq:steady-state-angle-rate}
\omega_\tx{ss}(j_\tx{ss},v_x) = \frac{j_\tx{ss}}{l}\left(\left(\frac{l}{v_x}\right)^2  + \frac{m}{2}\left(\frac{l_\tx{r}}{c_\tx{f}} - \frac{l_\tx{f}}{c_\tx{r}} \right)\right).
\end{align}
\end{Lemma}
\begin{proof}
The steady-state angle in \eqref{eq:steady-state-angle} follows directly from \eqref{eq:dynamic-bicycle} by solving for ${\ddot\psi=0}$ and $\dot v_\tx{s}=0$. The steady-state rate in \eqref{eq:steady-state-angle-rate} can be obtained by differentiating \eqref{eq:steady-state-angle}.
\end{proof}

\begin{figure}[t]
    \centering
	\subfigure[Steady-state lateral acceleration for a constant steering angle $\delta$.]{%
        \includegraphics{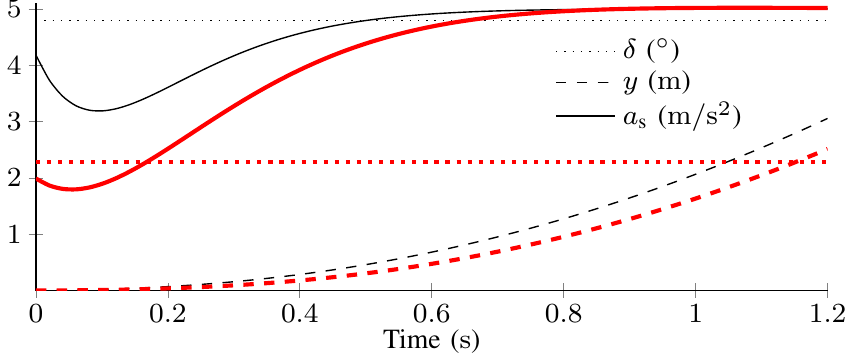}
	    \label{fig:ss_acc}
    }
    \subfigure[Steady-state lateral jerk for a constant steering rate $\omega$.]{%
     \includegraphics{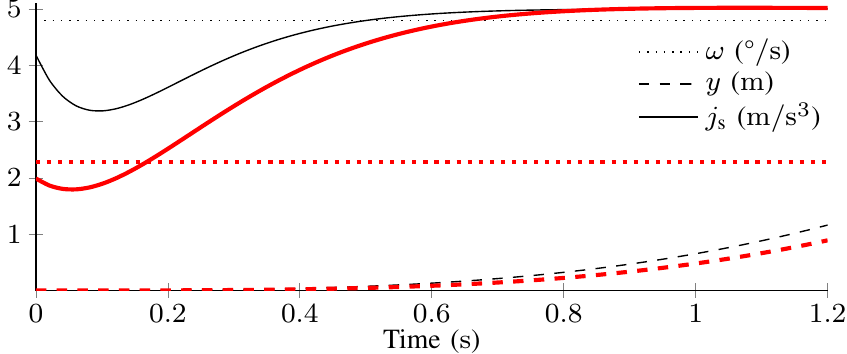}
	    \label{fig:ss_jerk}
    }
\caption{Illustration of lateral motion under constant steering angle and rate, respectively. The thin black lines depict the lateral motion (i.e., $\delta$, $y$, and $a_\tx{s}$) when longitudinal speed is \SI{50}{km/h}, and the thick red lines depict the lateral motion (i.e., $\delta$, $y$, and $a_\tx{s}$) when longitudinal speed is \SI{80}{km/h}. The initial lateral state $\mathbf x_\tx{s}$ is set to zero, except for the steering angle in (a), which is set to a value that results in a steady acceleration of \SI{5}{m/s^2}. It can be seen that steady values are reached in less than \SI{1}{s} for the different values of longitudinal speed.}
\label{fig:ss_acc_jerk}
\end{figure}

The idea is to limit the steady-state acceleration \eqref{eq:steady-state-angle} and jerk  \eqref{eq:steady-state-angle-rate} to the maximum lateral acceleration and jerk defined by the AJ profile; i.e., $a_\tx{ss}=a_\tx{smax}$ and $j_\tx{ss}=j_\tx{smax}$. These values together with the longitudinal vehicle speed, will provide the maximum steering angle and rate that can be used in the driver steering SASR profile. Fig.~\ref{fig:ss_acc_jerk} illustrates the steady-state acceleration and jerk when steering angle and rate,  respectively, are kept constant.

The steering angle and rate also need  be saturated to the physical vehicle limits, $\delta_\tx{Vmax}$ and $\omega_\tx{Vmax}$; further, they may be influenced by the road friction $\mu$. Vehicle operation within the linear tire region can be enforced by bounding tire forces to lie within a friction ellipse \cite[pp.~427-428]{rajamani06}. The area of the friction ellipse is proportional to the vertical force $\mu m g$, where $g$ is gravitational acceleration. Under the assumption of zero longitudinal slip during the steering manoeuvre, this bound is directly translated as a constraint on the lateral tire forces. The constraints can be formulated as
\begin{subequations} \label{eq:frictiion-elipse}
\begin{align}
	2c_\tx{f}|\alpha_\tx{f}| &= 2c_\tx{f}\left|\delta - \frac{l_\tx{f}\dot\psi + v_\tx{s}}{v_x}\right| \leq \mu m g\\
	2c_\tx{r}|\alpha_\tx{r}| &= 2 c_\tx{r} \left|\frac{l_\tx{r}\dot\psi - v_\tx{s}}{v_x}\right| \leq \mu m g
\end{align}
\end{subequations}
at the front and rear tire, respectively. With steady-state cornering, expressions for $v_\tx{s}$ and $\dot\psi$ can be derived as
\begin{subequations} \label{eq:ss-cornering-vy-dpsi}
	\begin{align}
		v_\tx{s} &= \frac{l_\tx{r} - \frac{m v_x^2 l_\tx{f}}{2c_\tx{r}l}}{l-\frac{mv_x^2}{2 l}\left( \frac{l_\tx{f}}{c_\tx{r}} - \frac{l_\tx{r}}{c_\tx{f}} \right)}v_x\delta\\
		\dot \psi &= \frac{1}{l-\frac{mv_x^2}{2 l}\left( \frac{l_\tx{f}}{c_\tx{r}} - \frac{l_\tx{r}}{c_\tx{f}} \right)}v_x\delta.
	\end{align}
\end{subequations}
Considering that for the studied manoeuvre the steady-state steering angle is nonnegative, these constraints can be compiled as
\begin{align}
	\begin{split} \label{eq:deltamax-friction}
	\delta &\leq \delta_{\tx{max}\mu}(\mu,v_x) \\
	&= \frac{\mu g}{\max(l_\tx{f},l_\tx{r})} \left( \left(\frac{l}{v_x}\right)^2 + \frac{m}{2}\left( \frac{l_\tx{r}}{c_\tx{f}} -\frac{l_\tx{f}}{c_\tx{r}} \right) \right). 
\end{split} 
\end{align}
Comparing \eqref{eq:deltamax-friction} with \eqref{eq:steady-state-angle} makes it evident that for a road friction coefficient
\begin{align}
    \mu \leq \frac{a_\tx{smax}}{g} \frac{\max(l_\tx{f},l_\tx{r})}{l_\tx{f}+l_\tx{r}}
\end{align}
the bound \eqref{eq:deltamax-friction} becomes more conservative than the bound limiting the steady-state acceleration to $a_\tx{smax}$. For example, for a maximum steady-state lateral acceleration of ${a_\tx{smax}=\SI{5}{m/s^2}}$, the friction threshold when \eqref{eq:deltamax-friction} becomes more conservative is about ${\mu\leq 0.285}$. 

\begin{figure}[t]
    \centering
	\subfigure[Limits on steering angle. The road friction limit becomes more conservative than the sensed acceleration limit for $\mu\leq 0.285$.]{%
    \includegraphics{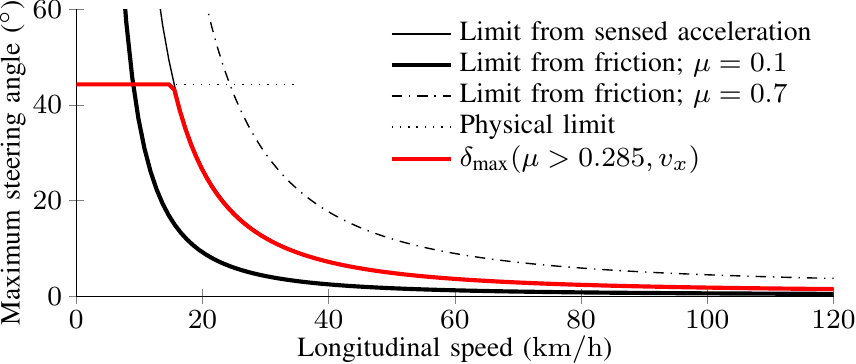}
	    \label{fig:steering_angle_vs_speed}
    }
    \subfigure[Limits on steering rate.]{%
    \includegraphics{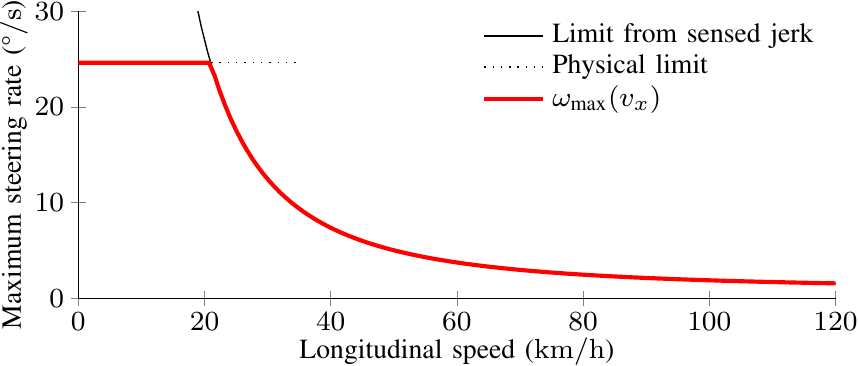}
	    \label{fig:steering_rate_vs_speed}
    }
\caption{Limits on steering angle and rate due to (a) sensed steady-state acceleration, road friction and physical vehicle limit and (b) sensed steady-state jerk, and physical vehicle limit.}
\label{fig:angle_rate_vs_speed}
\end{figure}

Finally, the maximum allowed steering angle and rate can be obtained as 
\begin{subequations} \label{eq:delta_omega_max_steering}
\begin{align}
	&\delta_\tx{max}(\mu,v_x) = \min(\delta_\tx{Vmax},\delta_\tx{ss}(a_\tx{smax}, v_x),  \delta_{\tx{max}\mu}(\mu,v_x))\\
	&\omega_\tx{max}(v_x) = \min(\omega_\tx{Vmax}, \omega_\tx{ss}(j_\tx{smax}, v_x)).
\end{align} 
\end{subequations}
An illustration of these limits is provided in Fig.~\ref{fig:angle_rate_vs_speed}. 

\begin{figure}[t]
	\centering
        \includegraphics{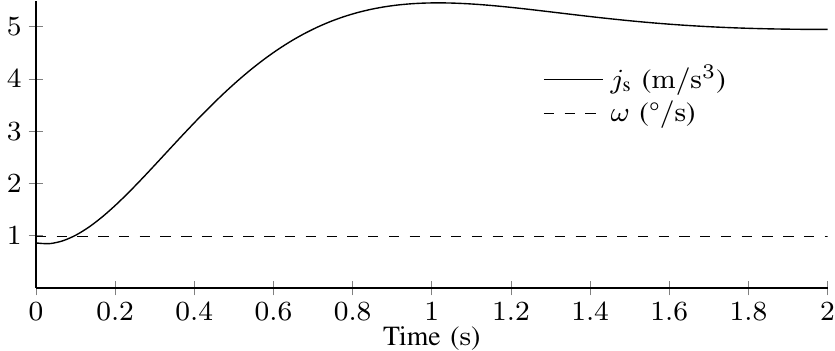}
    \caption{Steady-state lateral jerk for a constant steering rate $\omega$. The initial lateral state is set to zero and the longitudinal speed is \SI{180}{km/h}. 
    After reaching its maximum at about \SI{1}{s}, the lateral jerk eventually reaches a steady state at \SI{5}{m/s^3}.}
    \label{fig:ss_jerk_extremum}
\end{figure}

\begin{Remark}
The steady-state limits in \eqref{eq:steady-state-angle} and \eqref{eq:steady-state-angle-rate} do not guarantee that the maximum sensed acceleration and jerk ($a_\tx{smax}$ and $j_\tx{smax}$, respectively) are not exceeded before a steady-state condition is reached. In common steering scenarios, in which the initial vehicle state is 
\[v_\tx{s}(0)\approx\psi(0)\approx\dot\psi(0)\approx 0, \]
and the longitudinal speed is ${\SI{45}{km/h}\lessapprox v_x \lessapprox\SI{100}{km/h}}$, the maximum sensed acceleration and jerk often do occur at a steady-state, as illustrated in Fig.~\ref{fig:ss_acc_jerk}. It is possible to find conditions for which acceleration or jerk reach a higher extremum before reaching a steady state; see Fig.~\ref{fig:ss_jerk_extremum}. However, imposing constraints on these extrema is outside the scope of this paper.
\end{Remark}

\subsection{Collision avoidance by steering}

The time evolution of the steering states $\mathbf x_\tx{s}$ can be obtained as the solution of an  ordinary linear differential equation, as in \eqref{eq:ZOH}. For a constant steering action $\bar u_\tx{s}$,
a closed-form solution for the states evolution exists 
\begin{align} \label{eq:steering-dynamics-discrete}
    \mathbf x_\tx{s}(t) = A_{\tx{s}t}(v_x,t) \mathbf x_\tx{s}(0) + B_{\tx{s}t}(v_x,t) \bar u_\tx{s},
\end{align}
which provides the state vector $\mathbf x_\tx{s}(t)$ at any given time $t$. (Here, the matrices $A_{\tx{s}t}$ and $B_{\tx{s}t}$ are generally nonlinear functions of time.) The position of the front right vehicle corner at time $t_\tx{s}$ can be obtained by extracting the first element of the output vector
\begin{align}  \label{eq:yFR}
    y_\tx{FR}(t_\tx{s}) = H_y C_\tx{s}(v_x) \mathbf x_\tx{s}(t_\tx{s}) - \frac{W}{2}
\end{align}
where $H_y=\begin{bmatrix}1&0&0\end{bmatrix}$ is an output selector matrix. As described in Section~\ref{sec:problem_statement}, the minimum time $t_\tx{s}$ to avoid a collision can be obtained when the front right ego vehicle corner is just above the rear left corner of the leading vehicle; i.e., it holds, at the boundary, ${y_\tx{FR}(t_\tx{s}) = y_\tx{L} + y_\tx{margin}}$. From here, the sought steering time can be obtained as one of the roots of the function 
\begin{align} \label{eq:fcn_gs}
\begin{split}
    g_\tx{s}(t_\tx{s}) &= H_y C_\tx{s} \mathbf x_\tx{s}(t_\tx{s})- \frac{W}{2} - y_\tx{L} - y_\tx{margin} \\
    &= H_y C_\tx{s} \left( A_{\tx{s}t}(t_\tx{s}) \mathbf x_\tx{s}(0) + B_{\tx{s}t}(t_\tx{s}) \bar u_\tx{s} \right)\\
    &- \frac{W}{2} - y_\tx{L} - y_\tx{margin} = 0
\end{split}
\end{align}
(where the dependence on  inputs other than $t_\tx{s}$ is not shown explicitly, for didactic reasons). 

\begin{Conjecture}
A closed-form expression for the roots of $g_\tx{s}$, in the form ${t_\tx{s} = f_\tx{s}(\mathbf x_\tx{s}(0), \bar u_\tx{s}, v_x, y_\tx{L}, \mathbf p)}$, does not exist, in general.
\end{Conjecture}

A detailed proof of the conjecture would require careful investigation of the possible inputs to $f_\tx{s}$. Even if a closed-form solution exists for a set of realistic inputs, the solution would generally be too complex and computationally more expensive than using a numerical approximation calculated using a root-finding algorithm. Our analysis shows that even computing the system matrices $A_{\tx{s}t}$ and $B_{\tx{s}t}$ from a closed-form analytic expression is computationally more expensive than computing them numerically. In fact, Matlab is not able to symbolically evaluate
\begin{align*}
    \mathbf x_\tx{s}(t) = e^{A_\tx{s}(v_x) t} \mathbf x_\tx{s}(0) + \int_0^{t} e^{A_\tx{s}(v_x)(t-\tau)} B_\tx{s}(v_x) \bar u_\tx{s}\tx{d}\tau.
\end{align*}
Instead, the solution is investigated for the equivalent autonomous system 
\begin{align*}
    \frac{\tx{d}}{\tx{d}t}\begin{bmatrix} \mathbf x_\tx{s}\\ \bar u_\tx{s} \end{bmatrix} = A_\tx{se}(v_x) \begin{bmatrix} \mathbf x_\tx{s}\\ \bar u_\tx{s} \end{bmatrix}, \quad A_\tx{se}(v_x) = \begin{bmatrix} A_\tx{s}(v_x) & B_\tx{s}(v_x)\\0 & 0\end{bmatrix}.
\end{align*}
The autonomous system can be solved as 
\begin{align} \label{eq:xs_t}
     \begin{bmatrix} \mathbf x_\tx{s}(t) \\ \bar u_\tx{s}\end{bmatrix} = e^{A_\tx{se} t}\begin{bmatrix} \mathbf x_\tx{s}(0) \\ \bar u_\tx{s}\end{bmatrix} = \begin{bmatrix} A_{\tx{s}t} & B_{\tx{s}t}\\ 0 &  1\end{bmatrix} \begin{bmatrix} \mathbf x_\tx{s}(0) \\ \bar u_\tx{s}\end{bmatrix}, 
\end{align}
showing that the matrices $A_{\tx{s}t}$ and $B_{\tx{s}t}$ can be obtained if the matrix exponent $e^{A_\tx{se} t}$ is evaluated. To compute the matrix exponent, we first note that the matrix $A_\tx{se}$ is not diagonalizable. It has, in fact, four zero eigenvalues and one complex conjugate pair. One way to proceed is by computing its normal Jordan form $J(v_x)=P^{-1}(v_x)A_\tx{se}(v_x)P(v_x)$, where $P$ is invertible and $J$ is upper triangular. From Lemma~\ref{lemma:jordan} in Appendix~\ref{sec:lemmas}, it follows that
\begin{align}
    e^{A_\tx{se} t} = Pe^{Jt}P^{-1}
\end{align}
where the Jordan matrix exponent resolves to
\begin{align*}
    &e^{Jt} = \begin{bmatrix}
    1 & t & \frac{t^2}{2} & \frac{t^3}{6} &  & \\
     & 1 & t & \frac{t^2}{2} &  & \\   
     &  & 1 & t &  & \\
     &  &  & 1 &  & \\
     &  &  &  & e^{-t\frac{p_1 + p_5 - h(v_x)}{2v_x}} & \\
     &  &  &  &  & e^{-t\frac{p_1 + p_5 + h(v_x)}{2v_x}}
    \end{bmatrix}\\
    &h(v_x) =\sqrt{(p_1 - p_5)^2 + 4p_4(p_2 - v_x^2)}.
\end{align*}
From the matrix exponent, which involves a third order polynomial and exponential functions in $t$, it may be concluded that the function $g_\tx{s}$ may also include these terms and their cross-products. This conclusion supports the conjecture that a general closed-form solution is not  available. 

\subsection{Numerical solution for the steering time}

A plethora of numerical algorithms can be applied to find the roots of $g_\tx{s}$. Because computational efficiency is one of the main requirements, this section focusses on the Newton-Raphson and Halley’s methods.

The Newton-Raphson method is a powerful way to provide a local quadratic convergence to the function roots. It is often the method of choice for functions whose derivative can be evaluated efficiently, and the functions are continuous and the derivative is nonzero in the neighbourhood of a root  \cite[pp.~456-461]{Press2007}. The method is iterative; starting from an initial guess, the value for the steering time is updated as
\begin{align} \label{eq:t_k1}
    t_{k+1} = t_k - \tx{step} \frac{g_\tx{s}(t_k)}{\dot g_\tx{s}(t_k)}
\end{align}
where $\tx{step}\in(0, 1]$ and the derivative is computed as
\begin{align} \label{eq:dot_gs}
\begin{split}
    \dot g_\tx{s}(t) &= H_y C_\tx{s}(v_x) \dot{\mathbf x}_\tx{s}(t)\\
    &= H_y C_\tx{s}(v_x)(A_\tx{s}(v_x) \mathbf x_\tx{s}(t) + B_\tx{s}(v_x)\bar u_\tx{s}).
    \end{split}
\end{align}

Halley's method is another root-finding method    \cite[p.~463]{Press2007}, in which the steering time is updated as
\begin{align}
    t_{k+1} = t_k - \tx{step} \frac{g_\tx{s}(t_k)}{\dot g_\tx{s}(t_k) - \frac{g_\tx{s}(t_k) \ddot g_\tx{s}(t_k)}{2\dot g_\tx{s}(t_k)}}
\end{align}
with the second derivative computed as
\begin{align} \label{eq:ddot_gs}
\begin{split}
    \ddot g_\tx{s}(t) &= H_y C_\tx{s}(v_x) \ddot{\mathbf x}_\tx{s}(t) = H_y C_\tx{s}(v_x) A_\tx{s}(v_x) \dot{\mathbf x}_\tx{s}(t) \\
    &= H_y C_\tx{s}(v_x)A_\tx{s}(v_x)(A_\tx{s}(v_x) \mathbf x_\tx{s}(t) + B_\tx{s}(v_x)\bar u_\tx{s}).
\end{split}
\end{align}
The method provides a local cubic convergence, but each step in its iteration is computationally more expensive than the steps in the Newton-Raphson method, so the former  is not commonly used. The studied problem, however, is well suited for  Halley's method, because the problem is one-dimensional and evaluation and inversion of the function derivatives are cheap. 

\begin{figure}[t]
	\centering
	\includegraphics{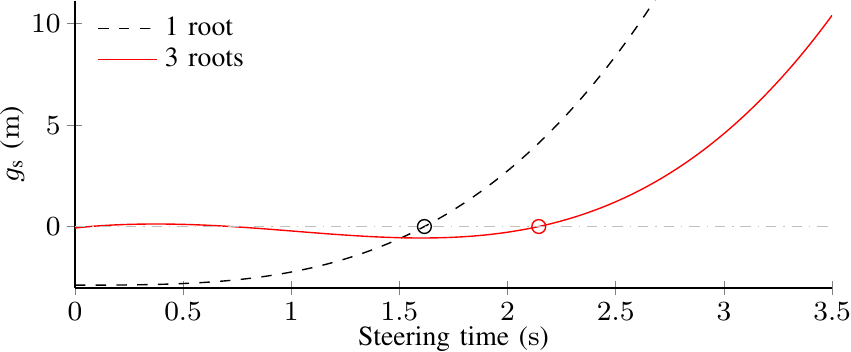}
    \caption{Illustration of the function $g_\tx{s}$ for $v_x=\SI{80}{km/h}$ and different initial vehicle states. The function with one root has an initial state of zero, while the other function is initialised at $\mathbf x_\tx{s}(0)=[\SI{2.75}{m},\SI{2}{\degree},\SI{0.5}{m/s},\SI{0}{\degree/s},\SI{-2}{\degree}]$. The circles indicate the roots with the greatest steering time, $t_\tx{s}$ (i.e., the time from the start of steering until the lateral position $y_\tx{FR}(t_\tx{s})$, given in~\eqref{eq:steering-time}, is reached).}
    \label{fig:function_roots}
\end{figure}

It remains to investigate whether the methods may fail to solve the problem. In typical driving scenarios with an initial vehicle state
\[v_\tx{s}(0)\approx\psi(0)\approx\dot\psi(0)\approx 0, \]
the function $g_\tx{s}$ is a monotonic convex function. However, for different initial conditions the function may have multiple roots; see, for example, Fig.~\ref{fig:function_roots}. 
Out of these, the root with the longest steering time, $t_\tx{s}$ in \eqref{eq:steering-time}, is the correct solution. Moreover, the function may have extrema where $\dot g_\tx{s}=0$, which could cause singularities. A simple fix for this issue is to initialise the methods with a large steering time and let the algorithms converge from the right, where $g_\tx{s}$ is locally convex and monotonic. 
\begin{figure}[t]
	\centering
	\includegraphics{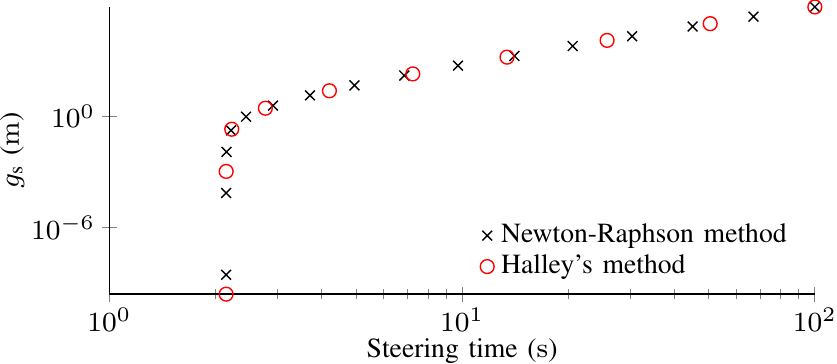}
    \caption{Convergence of the Newton-Raphson and Halley's methods with ${\tx{step}=1}$ applied to the function with three roots in Fig.~\ref{fig:function_roots}. The algorithms are initialised with $t_0=\SI{100}{s}$ and stop when $\vert g_\tx{s}(t_k)\vert<\SI{e-6}{m}$. The Newton-Raphson and Halley's methods converge at 16 and 10 iterations, respectively, with both achieving much higher accuracy than requested. Their computation times are  \SI{0.53}{ms} and \SI{0.34}{ms}, respectively, on a standard PC with a boost CPU frequency of \SI{4.2}{GHz}. Both algorithms converge to the correct steering intervention time to avoid a crash  (about \SI{2.14}{s}).}
    \label{fig:newton_halley}
\end{figure}

This  property (local convexity and monotonicity) allows full convergence steps to be taken; i.e., $\tx{step}=1$ from \eqref{eq:t_k1}. Furthermore, note that although $C_\tx{s}$ is a function of $v_x$, the product $H_y C_\tx{s}$ is not, which can be seen directly in the first row of the $C_\tx{s}$ matrix in Table~\ref{TT:ModelMatrices}, Appendix~\ref{sec:different-models}. Hence, this product needs only to be computed once. The matrix $A_\tx{s}$ needs to be evaluated only when the longitudinal speed changes, but it remains constant during the iterations of the root-finding algorithms. The most computationally demanding part is evaluating $\mathbf x_\tx{s}(t)$ from \eqref{eq:xs_t}, which has to be performed in each iteration. In this study, to compute the matrix exponent, we used the $\mathrm{expm}$ function in Matlab. (For other implementations see, for example, \cite{moler03}.)

The algorithms are fast even when the initial guess is several orders of magnitude greater than the solution; see Fig.~\ref{fig:newton_halley}. 

\subsection{Algorithm for collision avoidance by steering}

The root-finding methods described in the previous section essentially provide an implicit implementation of the function $f_\tx{s}$. As a result, if the maximum steering rate is used as the steering action, the steering time 
\begin{align} \label{eq:steering_time_jerk}
    t_\tx{sj} = f_\tx{s}(\mathbf x_\tx{s}(0), \omega_\tx{max}, v_x, y_\tx{L}, \mathbf p)
\end{align}
is obtained. Next, the time
\begin{align} \label{eq:steering_time_acc}
    t_\tx{sa} = \frac{\delta_\tx{max}(\mu,v_x)-\delta(0)}{\omega_\tx{max}(v_x)}
\end{align}
when the maximum steering angle is reached can be computed. If $t_\tx{sa}\geq t_\tx{sj}$, then the steering time is obtained as $t_\tx{s}=t_\tx{sj}$. Otherwise, \eqref{eq:xs_t} needs to be computed first until time $t_\tx{sa}$ with $\bar u_\tx{s}=\omega_\tx{max}$, to obtain the state vector $\mathbf x_\tx{s}(t_\tx{sa})$. Taking $\mathbf x_\tx{s}(t_\tx{sa})$ as the initial state, \eqref{eq:xs_t} needs to be computed one more time with $\bar u_\tx{s}=0$, and the steering time can be obtained as
\begin{align}
    t_\tx{s} = t_\tx{sa} + f_\tx{s}(\mathbf x_\tx{s}(t_\tx{sa}), 0, v_x, y_\tx{L}, \mathbf p).
\end{align}

The final step is computing the longitudinal position $x_\tx{s}(t_\tx{s})$ from \eqref{eq:dot_xs} and the relative longitudinal distance $\Delta x_\tx{s}(t_\tx{s})$ from \eqref{eq:delta_xs}. Notice that \eqref{eq:dot_xs} involves the integration of a nonlinear function that may not have a closed-form solution. Here, we implement an approximate numerical integration. 

The algorithm for avoiding collision by steering can be summarised as in Algorithm~\ref{alg:fy}. 
\begin{algorithm}[tb]
\SetAlgoLined
\DontPrintSemicolon
  \textbf{Inputs}: $\mathbf x_\tx{s}(0)$, $\mathbf x_\tx{b}(0)$, $x_\tx{L}(0)$, $v_\tx{L}$, parameters\;
  Compute $\delta_\tx{max}$, $\omega_\tx{max}$ and $t_\tx{sa}$ from \eqref{eq:delta_omega_max_steering} and \eqref{eq:steering_time_acc}\;
  Using a control action $\bar u_\tx{s}=\omega_\tx{max}$ compute $t_\tx{sj}$ from \eqref{eq:steering_time_jerk}\;
  \If{$t_\tx{sj}\leq 0$}{There is no risk of collision\\Exit}
  \eIf{$t_\tx{sa}\geq t_\tx{sj}$}{
    $t_\tx{s}=t_\tx{sj}$\;
    $\mathbf x_\tx{s}(t_\tx{s}) = A_{\tx{s}t}(v_x,t_\tx{s})\mathbf x_\tx{s}(0) + B_{\tx{s}t}(v_x,t_\tx{s}) \omega_\tx{max}$\;
    }  
    {
    $\mathbf x_\tx{s}(t_\tx{sa}) = A_{\tx{s}t}(v_x,t_\tx{sa})\mathbf x_\tx{s}(0) + B_{\tx{s}t}(v_x,t_\tx{sa}) \omega_\tx{max}$\;
    $t_\tx{s} = t_\tx{sa} +  f_\tx{s}(\mathbf x_\tx{s}(t_\tx{sa}), 0, v_x, y_\tx{L}, \mathbf p)$\;
    $\mathbf x_\tx{s}(t_\tx{s}) = A_{\tx{s}t}(v_x,t_\tx{s}-t_\tx{sa})\mathbf x_\tx{s}(t_\tx{sa})$\;
    }
    Discretise the time $\mathcal T_\tx{d}\subset[0,t_\tx{s}]$, obtain the evolution of the lateral states from \eqref{eq:xs_t} for all instants in $\mathcal T_\tx{d}$, and compute $\Delta x_\tx{s}(t_\tx{s})$ from \eqref{eq:dot_xs} and \eqref{eq:delta_xs}\;
    \eIf{$-\Delta x_\tx{s}(t_\tx{s})\geq x_\tx{margin}$}{
    Collision can be avoided by steering\;
    }
    {Collision cannot be avoided by steering\;
    }
 \caption{Collision avoidance by steering.}
 \label{alg:fy}
\end{algorithm}

\subsection{Simplified algorithm for collision avoidance by steering}

A drawback of Algorithm~\ref{alg:fy} is the computation of the longitudinal steering distance $x_\tx{s}(t_\tx{s})$. As mentioned earlier, the integration in \eqref{eq:dot_xs} is performed numerically, which is an approximate solution, as its accuracy depends on the choice of integration method and the length of the sampling interval. A shorter sampling interval provides a more accurate solution, but it increases computation time, since \eqref{eq:xs_t} needs to be evaluated at each sample. In order to reduce computation time, we propose approximating the longitudinal distance travelled during the steering manoeuvre with $x_\tx{s}(t_\tx{s}) \approx x(0)+v_xt_\tx{s}$, where $x(0)=x_\tx{s}(0)=x_\tx{b}(0)$ is the initial longitudinal position, which is identical whether steering or braking is performed. In effect, discretisation of the time interval $[0, t_\tx{s}]$ is not needed. 
The simplified algorithm for avoiding collision by steering can be summarised as in Algorithm~\ref{alg:fy_simplified}.

\begin{algorithm}[tb]
\SetAlgoLined
\DontPrintSemicolon
  Everything is the same as Algorithm~\ref{alg:fy} except step 16 is replaced by $x_\tx{s}(t_\tx{s}) = x(0)+v_xt_\tx{s}$.
 \caption{Simplified collision avoidance by steering.}
 \label{alg:fy_simplified}
\end{algorithm}

Another consequence of this approximation is that the steering time may now be computed as 
\begin{align}\label{eq:t_s_aprox}
    t_\tx{s} = \frac{-x_\tx{margin}+x_\tx{L}(0)-x(0)-L_\tx{f}}{v_x-v_L}
\end{align}
using forward simulation and without the need for either Newton-Raphson or Halley's method. After steering time is obtained, collision avoidance can be checked by investigating the lateral displacement of the ego vehicle's front right corner. 
This check is useful for online real-time usage, when the algorithm will always run in the background and only issue an intervention when a lateral displacement satisfies condition \eqref{eq:lat-safety-margin} approximately with equality. However, for offline assessment where intervention time is computed backwards, this will not be particularly faster than Algorithm~\ref{alg:fy_simplified} since it will require iteratively simulating system \eqref{eq:dynamic-bicycle-lateral} for $t_\tx{s}$ seconds  until the previous condition is satisfied. 
The simplified algorithm for avoiding the collision by steering using forward simulation can be summarised as in Algorithm~\ref{alg:fy2}.
\begin{algorithm}[tb]
\SetAlgoLined
\DontPrintSemicolon
  \textbf{Inputs}: $\mathbf x_\tx{s}(0)$, $\mathbf x_\tx{b}(0)$, $x_\tx{L}(0)$, $v_\tx{L}$, parameters\;
  Compute $\delta_\tx{max}$, $\omega_\tx{max}$ and $t_\tx{sa}$ from \eqref{eq:delta_omega_max_steering} and \eqref{eq:steering_time_acc}\;
  Compute $t_\tx{s}=\frac{-x_\tx{margin}+x_\tx{L}(0)-x(0)-L_\tx{f}}{v_x-v_L}$\;
  \eIf{$t_\tx{sa}\geq t_\tx{s}$}{
    $\mathbf x_\tx{s}(t_\tx{s}) = A_{\tx{s}t}(v_x,t_\tx{s})\mathbf x_\tx{s}(0) + B_{\tx{s}t}(v_x,t_\tx{s}) \omega_\tx{max}$\;
    }  
    {
    $\mathbf x_\tx{s}(t_\tx{sa}) = A_{\tx{s}t}(v_x,t_\tx{sa})\mathbf x_\tx{s}(0) + B_{\tx{s}t}(v_x,t_\tx{sa}) \omega_\tx{max}$\;
    $\mathbf x_\tx{s}(t_\tx{s}) = A_{\tx{s}t}(v_x,t_\tx{s}-t_\tx{sa})\mathbf x_\tx{s}(t_\tx{sa})$\;
    }
   Compute $y_\tx{FR}(t_\tx{s})$ from \eqref{eq:yFR}\;
    \eIf{$y_\tx{FR}(t_\tx{s})\geq y_\tx{L} + y_\tx{margin}$}{
    Collision can be avoided by steering\;
    }
    {Collision cannot be avoided by steering\;
    }
 \caption{Simplified collision avoidance by steering using forward simulation.}
 \label{alg:fy2}
 \end{algorithm}

\section{Benchmark models for lateral motion}\label{sec:benchmarkModels}

The steering dynamic bicycle model used in this study is benchmarked with simplified vehicle models.

\subsection{Steady-state-cornering-inspired model}

The SSCM has previously been applied in \cite{Brannstrom2010}. The model uses three states 
\begin{align}
	\mathbf{x}_\tx{s}&=\begin{bmatrix} y & \psi & \delta \end{bmatrix}^\top
\end{align}
to model its lateral motion, while the output vector is identical to that in \eqref{eq:vectors_bicycle_model}. Its output and state equations have already been derived in \eqref{eq:steady-state-angle}, \eqref{eq:steady-state-angle-rate} and \eqref{eq:ss-cornering-vy-dpsi}. A complete description is summarised in Table~\ref{TT:ModelMatrices} in Appendix~\ref{sec:different-models}.

Algorithm~\ref{alg:fy} is used with the SSCM; however, the discretisation is not needed since the analytical integral is devised for calculating $x_\tx{s}(t_\tx{s})$ in Appendix~\ref{sec:different-models}. In this paper, we calculate the $x_\tx{s}(t_\tx{s})$ from \eqref{eq:dot_xs}, which is more accurate than the approach taken in \cite{Brannstrom2010} where $x_\tx{s}(t_\tx{s})$ is approximated with $v_\tx{x}t_\tx{s}$.

\subsection{Kinematic model}

The KM is yet another approximation of the DM \eqref{eq:dynamic-bicycle-lateral}. To the best of our knowledge, this model has not been used for collision avoidance, but it was worth investigating, since its complexity is between the SSCM and PMM.  The model assumes zero tire slip angles. Replacing this assumption in the following relations
\begin{subequations}
\begin{align}
	&\tan(\delta-\alpha_\tx{f})=\frac{v_\tx{s} + l_\tx{f} \dot \psi}{v_x}\approx \delta - \alpha_\tx{f}=\delta\\
	&\tan(\alpha_\tx{r})=-\frac{v_\tx{s} - l_\tx{r} \dot \psi}{v_x}\approx\alpha_\tx{r}=0
\end{align}
\end{subequations}
gives
\begin{align}
	\dot y &= v_x\left(\psi + \frac{l_\tx{r}}{l}\delta\right), \quad \dot\psi = \frac{v_x}{l}\delta.
\end{align}
The lateral acceleration and jerk can be expressed as
\begin{align}
	a_\tx{s} &= \frac{v_x}{l}\left( l_\tx{r}\omega + v_x\delta \right), \quad j_\tx{s} = \frac{v_x^2}{l}\omega.
\end{align}
The input, state, and output vectors of this model are identical to those in the SSCM, while a complete description of the system matrices is provided in Table~\ref{TT:ModelMatrices} in Appendix~\ref{sec:different-models}.

Algorithm~\ref{alg:fy} is used for collision avoidance by steering when using KM; however, similarly as with the SSCM, the discretisation step is not needed for this model since the analytical integral is devised for calculating $x_\tx{s}(t_\tx{s})$ in Appendix~\ref{sec:different-models}.

\subsection{Point mass model} \label{sec:PMM}
The PMM is simply described by
\begin{align}
	\dot y &= v_\tx{s}, \quad \dot v_\tx{s} = a_\tx{s}, \quad \dot a_\tx{s} = j_\tx{s}.
\end{align}
This model has previously been utilised in, for example, \cite{chandru17}. As this model does not possess signals for steering angle and rate, lateral jerk is  used directly as the control input. Its state vector is defined as 
\begin{align}
    \mathbf x_\tx{s}=\begin{bmatrix}y&v_\tx{s}&a_\tx{s}\end{bmatrix}, 
\end{align}
while its output vector is identical to the previous models. A complete description of the system matrices is provided in Table~\ref{TT:ModelMatrices} in Appendix~\ref{sec:different-models}.

For comparison, Algorithm~\ref{alg:fy} is used for collision avoidance by steering when using PMM; once again, however, the discretisation step is not needed since the analytical integral is devised for calculating $x_\tx{s}(t_\tx{s})$ in Appendix~\ref{sec:different-models}.

\section{Results} \label{sec:Results}
\begin{figure}[t]
\centering
\subfigure[$v_x=\SI{50}{km/h}$]{%
\includegraphics{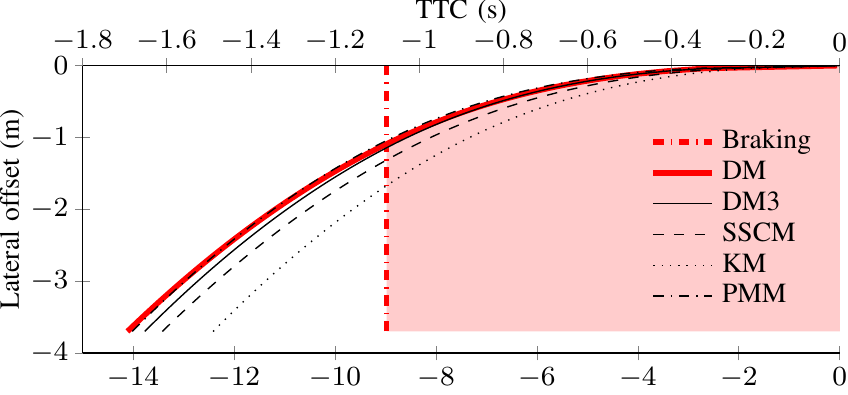}
\label{fig:M_vx50}
}
\subfigure[$v_x=\SI{70}{km/h}$]{%
\includegraphics{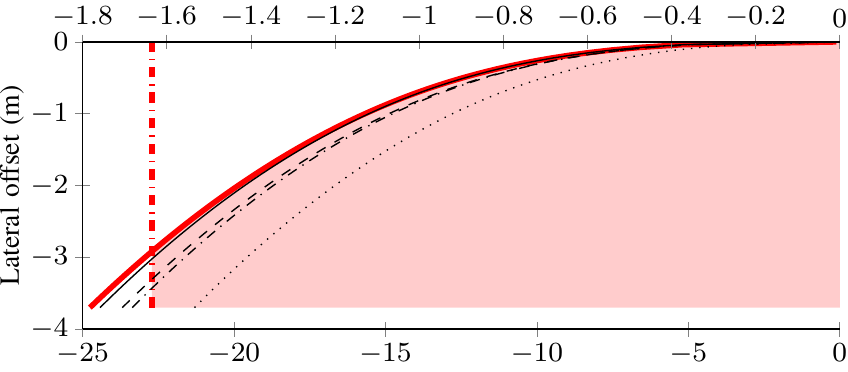}
\label{fig:M_vx70}
}
\subfigure[$v_x=\SI{90}{km/h}$]{%
\includegraphics{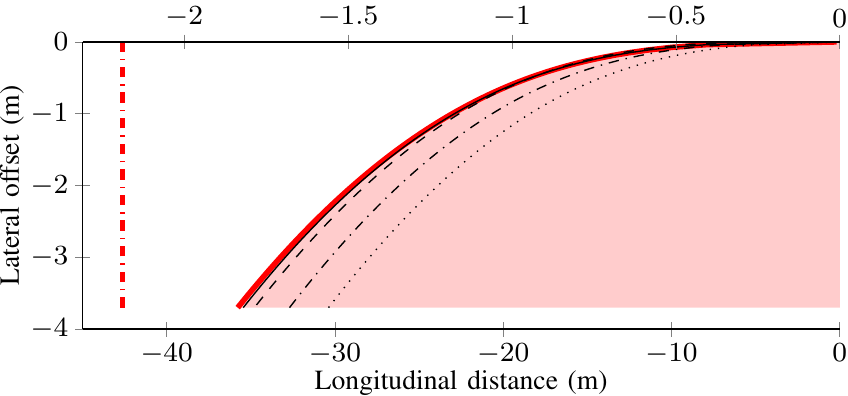}
\label{fig:M_vx90}
}
\caption{Critical zones for different vehicle models and different $v_x$, with initial conditions $\psi(0)=\SI{0}{\degree}$, $v_\tx{s}(0)=\SI{0}{m/s}$, $\dot\psi(0)=\SI{0}{\degree/s}$, and $\delta(0)=\SI{0}{\degree}$. The shaded red area shows the critical zone of the dynamic model (DM).}
\label{fig:M_vx}
\end{figure}

\begin{figure}[t]
\centering
\subfigure[$\psi(0)=\SI{-2}{\degree}$]{%
\includegraphics{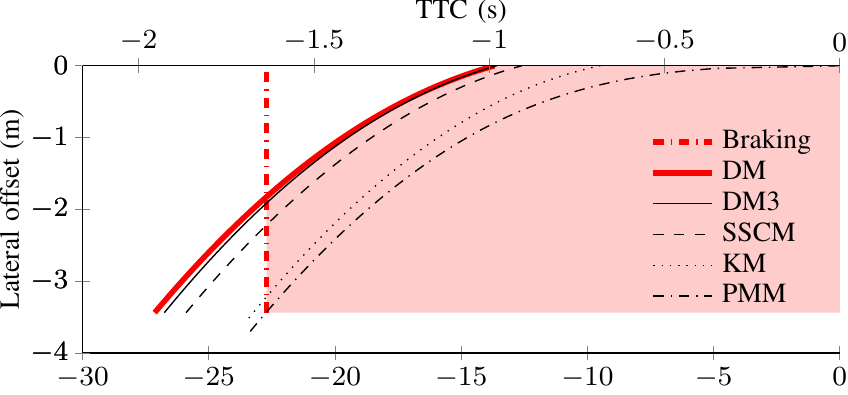}
\label{fig:M_vx70_psi-5}
}
\subfigure[$\psi(0)=\SI{2}{\degree}$]{%
\includegraphics{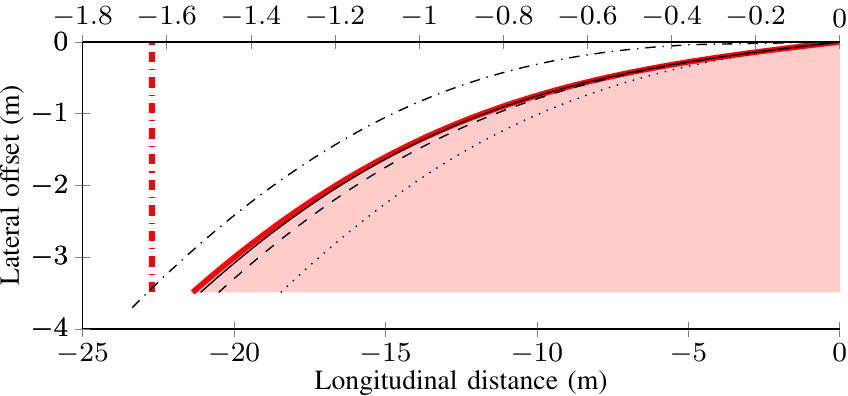}
\label{fig:M_vx70_psi5}
}
\caption{Critical zones for different vehicle models with $v_x=\SI{70}{km/h}$ and initial conditions ${v_\tx{s}(0)=\SI{0}{m/s}}$, $\dot\psi(0)=\SI{0}{\degree/s}$, and $\delta(0)=\SI{0}{\degree}$. The shaded red area shows the critical zone of the DM.}
\label{fig:M_psi}
\end{figure}

\begin{figure}[t]
\centering
\subfigure[$v_\tx{s}(0)=\SI{-0.5}{m/s}$]{%
\includegraphics{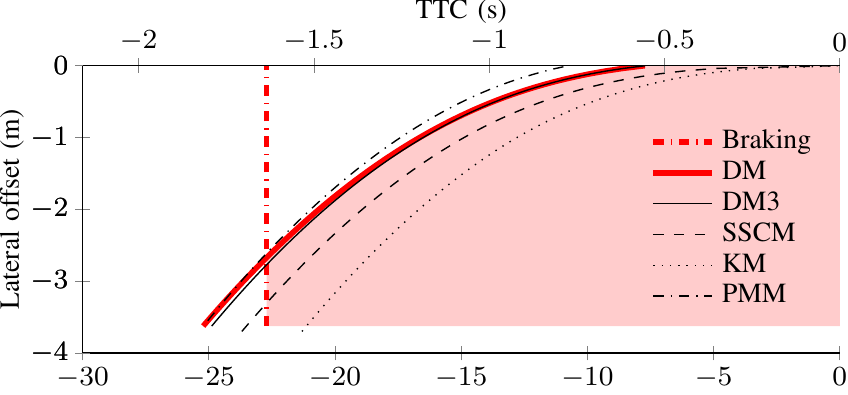}
\label{fig:M_vx70_vy-5}
}
\subfigure[$v_\tx{s}(0)=\SI{0.5}{m/s}$]{%
\includegraphics{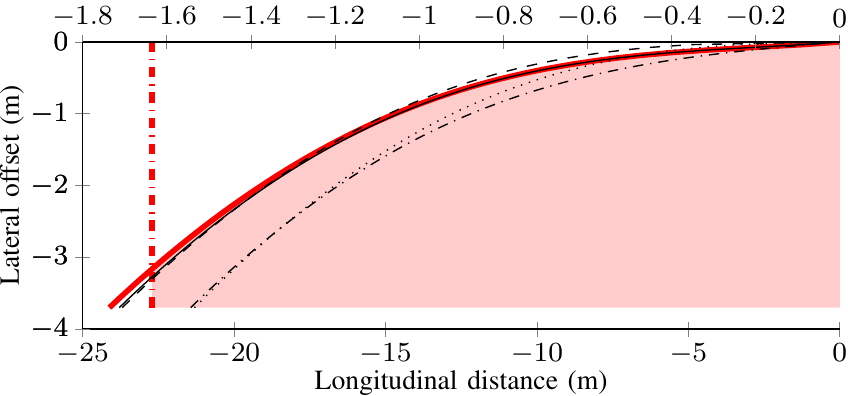}
\label{fig:M_vx70_vy5}
}
\caption{Critical zones for different vehicle models with $v_x=\SI{70}{km/h}$ and initial conditions ${\psi(0)=\SI{0}{\degree}}$, $\dot\psi(0)=\SI{0}{\degree/s}$, and $\delta(0)=\SI{0}{\degree}$. The shaded red area shows the critical zone of the DM.}
\label{fig:M_vy}
\end{figure}

\begin{figure}[t]
\centering
\subfigure[$\dot{\psi}(0)=\SI{5}{\degree/s}$]{%
\includegraphics{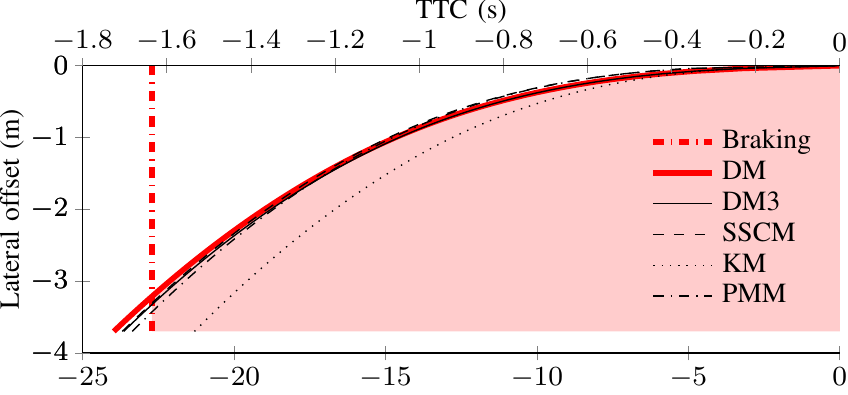}
\label{fig:M_vx70_dotpsi5}
}
\subfigure[$\dot{\psi}(0)=\SI{-5}{\degree/s}$]{%
\includegraphics{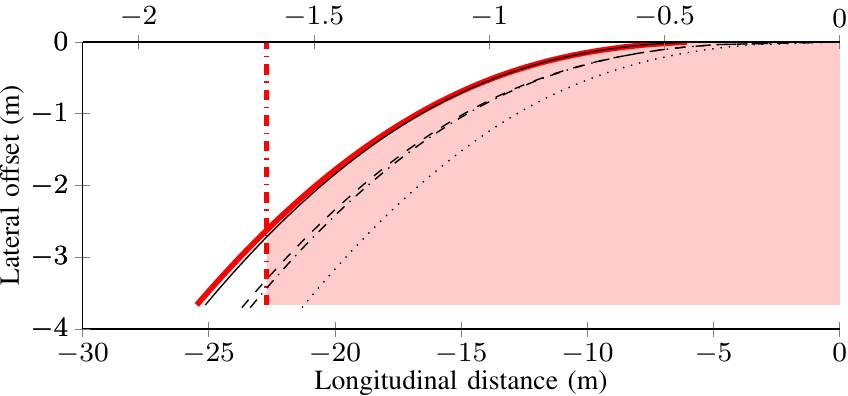}
\label{fig:M_vx70_dotpsi-5}
}
\caption{Critical zones for different vehicle models with $v_x=\SI{70}{km/h}$ and initial conditions ${\psi(0)=\SI{0}{\degree}}$, $v_\tx{s}(0)=\SI{0}{m/s}$, and $\delta(0)=\SI{0}{\degree}$. The shaded red area shows the critical zone of the DM.}
\label{fig:M_dotPsi}
\end{figure}

\begin{figure}[t]
\centering
\subfigure[$\delta(0)=\SI{-2}{\degree}$]{%
\includegraphics{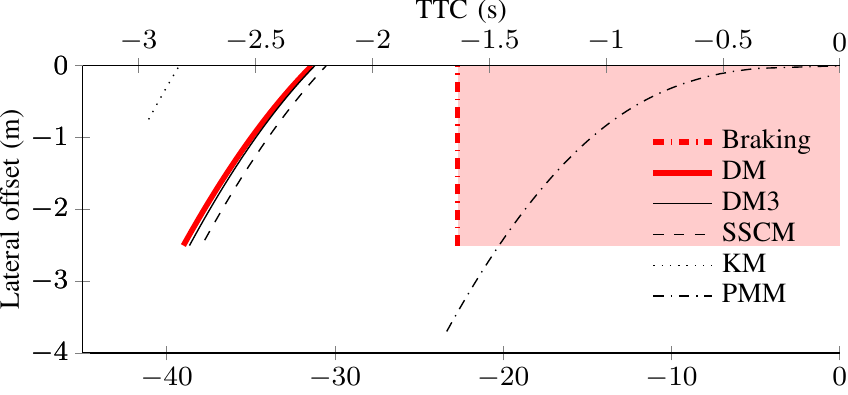}
\label{fig:M_delta-2}
}
\subfigure[$\delta(0)=\SI{2}{\degree}$]{%
\includegraphics{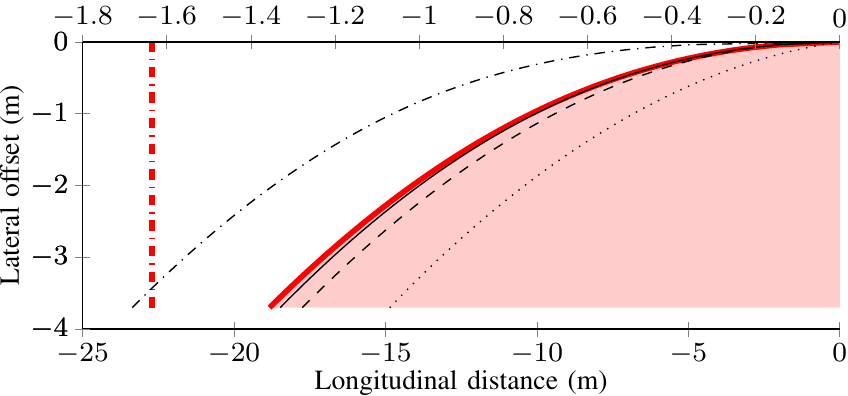}
\label{fig:M_delta2}
}
\caption{Critical zones for different vehicle models with $v_x=\SI{70}{km/h}$ and initial conditions ${\psi(0)=\SI{0}{\degree}}$, $v_\tx{s}(0)=\SI{0}{m/s}$, and $\dot\psi(0)=\SI{0}{\degree/s}$. The shaded red area shows the critical zone of the DM.}
\label{fig:M_delta}
\end{figure}

In this section, the steering bicycle DM is benchmarked with simplified vehicle models in simulated overtaking scenarios for different initial conditions: 
$v_x\in\{50, 70, 90\}$\,\SI{}{km/h}, $\psi\in\{-2,2\}\,\SI{}{\degree}$, $v_\tx{s}\in\{-0.5,0.5\}\,\SI{}{m/s}$, $\dot{\psi}\in\{-5,5\}\,\SI{}{\degree/s}$, and $\delta\in\{-2,2\}\,\SI{}{\degree}$, for the models having the respective states. The initial conditions for $v_x$ are typical for rural and highway driving;  the values for $\psi$, $v_\tx{s}$, and $\delta$ have been recorded in a test-track experiment \cite{Rasch2020, Rasch2020a}, while the values for $\dot{\psi}$ seem reasonable according to \cite{Scanlon:2015, Dingus:2006, Kusano:2013}. In each simulation, a range of lateral offsets $[-3.7, 0]\,\SI{}{m}$ is used. One of the values, either $v_x$, $\psi$, $v_\tx{s}$, $\dot{\psi}$ or  $\delta$, is changed at a time while the others are kept constant. The lateral offset of \SI{-3.7}{m} corresponds to a typical lane width \cite{AASHTO2011, ERSO2018}; the leading vehicle would be positioned near the left lane mark and the ego vehicle near the right lane mark. A speed of a leading road user of 
$v_\tx{L}=\SI{20}{km/h}$, an average cyclist speed  \cite{Kassim:2020}, is used in the calculations. We chose this scenario (a driver overtaking a cyclist) because it is very common. In fact, due to the large speed differences, it accounts for 10–\SI{49}{\%} of all fatal crashes between cars and bicyclists and 7–29\% of all crashes with serious injuries \cite{ITARDA:2018, Wisch:2017, Fredriksson:2014, Uittenbogaard:2016, MacAlister:2015}.
 Driver comfort boundaries for normal driving behaviour are considered from previous research \cite{Brannstrom2010}, shown in 
\begin{table}[!tb]
\caption{Driver comfort boundaries and vehicle parameters}
\centering
{\footnotesize
\begin{tabular}{lll}
\hline
$a_\tx{bmin}=\SI{-5}{m/s^2}$ & $L=\SI{4.27}{m}$ & $L_\tx{f}=\SI{1.820}{m}$\\
$j_\tx{bmin}=\SI{-10}{m/s^3}$ & $W=\SI{1.78}{m}$ & $l_\tx{r}=\SI{1.550}{m}$\\
$a_\tx{smax}=\SI{5}{m/s^2}$  & $m=\SI{2000}{kg}$ & $l_\tx{f}=\SI{1.226}{m}$\\
$j_\tx{smax}=\SI{5}{m/s^3}$  & $c_\tx{r}=\SI{50000}{N}$ & $c_\tx{f}=\SI{50000}{N/rad}$\\
$\delta_\tx{Vmax}=\SI{44.30}{\degree}$ &  $\omega_\tx{Vmax}=\SI{24.61}{\degree/s}$  &  $I_z=\SI{3200}{kgm^2}$\\
\hline
\end{tabular}}
\label{TT:Limits}
\end{table}
Table~\ref{TT:Limits}, together with vehicle parameters. The comparison is made in terms of a critical zone. 
\begin{figure}[t]
\centering
\includegraphics{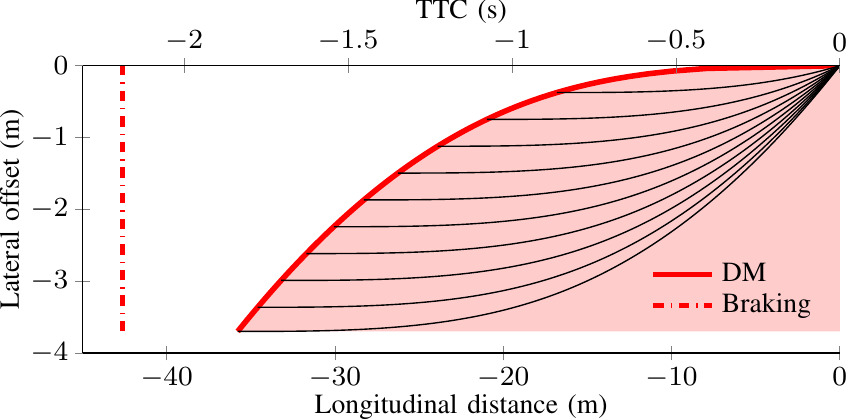}
\caption{Illustration of the critical zone of the dynamic model (DM) for ${v_x=\SI{90}{km/h}}$,  $v_\tx{L}=\SI{20}{km/h}$, and lateral offset in the interval $[-3.7, 0]$\,\SI{}{m}. The remaining initial conditions are $\psi(0)=\SI{0}{\degree}$, ${v_\tx{s}(0)=\SI{0}{m/s}}$, $\dot\psi(0)=\SI{0}{\degree/s}$, and $\delta(0)=\SI{0}{\degree}$. The thin black lines show the vehicle trajectories for ten initial, uniformly placed,  lateral positions within the given range, the thick red solid line shows the steering critical zone boundary, the shaded red area shows the critical zone, and the thick red dashed line shows the braking critical zone boundary.}
\label{fig:LDM_Explanation}
\end{figure}
The principle for obtaining the critical zone, described in Section~\ref{sec:Intro}, for the DM is illustrated in Fig.~\ref{fig:LDM_Explanation}, and follows in the remaining  Figs.~\ref{fig:M_vx}-\ref{fig:M_delta}. Specifically, the critical zone is the intersection between the braking and steering zones: the area in a specific state-space where a collision cannot be avoided, given the boundaries (e.g., for the comfort zone or performance). When Algorithm~\ref{alg:fx} is used to compute the time at which braking needs to be applied to avoid a crash, the longitudinal distance needed by the driver to avoid a collision by braking is obtained. This distance delineates the boundary for the braking critical zone.  As mentioned earlier, the same braking model is used for all the case studies, while the steering models may differ. Hence, the braking critical zone boundary is shown by a single dash-dotted line; see, for example, Fig.~\ref{fig:LDM_Explanation}. For braking distances to the right of the braking critical zone boundary,  drivers cannot avoid a collision by braking comfortably, but they  may still be able to avoid the crash by steering comfortably. The steering critical zone boundary is represented by the thick solid red line, computed for different initial lateral offsets between the lead vehicle and the ego vehicle. For each initial offset,  the trajectory associated with comfortable crash avoidance is shown as a thin solid black line in Fig.~\ref{fig:LDM_Explanation}: that is, these lines show the different crash avoidance trajectories that are possible with steering manoeuvres that reach the comfort zone boundaries. The steering intervention time is directly related to the steering critical zone boundary. For example, for lateral offsets of \SI{-3.7}{m} and \SI{-1.5}{m}, the longitudinal distances at which the steering must be initiated to avoid a crash are \SI{35.7}{m} and \SI{26.3}{m}, respectively. The driver cannot avoid the collision by steering for the positions below and to the right of the red line in Fig.~\ref{fig:LDM_Explanation} without exceeding the comfort zone boundaries. The x-axis shows the longitudinal distance as $\Delta x_\tx{s}+x_\tx{margin}$ and the y-axis shows the lateral offset as $y_\tx{FR}-y_\tx{L}-y_\tx{margin}$. The top x-axis shows TTC calculated as the ratio of the longitudinal distance between the ego and the leading vehicle and their relative speed. As the longitudinal speed is assumed to be constant, there is a linear relationship between distance and TTC.   

Figures~\ref{fig:M_vx}-\ref{fig:M_delta}  compare the critical zones of the four vehicle models when the zones are computed using Algorithm~\ref{alg:fy} (for the different vehicle models and different initial conditions). We also show the critical zone computed using Algorithm~\ref{alg:fy_simplified} for the dynamic model, which is hereafter referred to as DM3.

 The critical zones of the vehicle models for different $v_x$ are compared in Fig.~\ref{fig:M_vx}. It can be observed that the differences  in the steering distance required to avoid a collision increase with the lateral offset; as the lateral offset  decreases the difference decreases, regardless of the value of $v_x$. The critical zones for DM, DM3, and SSCM are similar, while that of KM exhibits the largest differences for all $v_x$. Interestingly,  despite the  low accuracy expected, the critical zone boundary for PMM is similar (with a maximum error of  \SI{0.1}{m}) to the DM for the lowest speed ($v_x=\SI{50}{km/h}$), and the largest difference is observed for the highest speed ($v_x=\SI{90}{km/h}$), see Figs.~\ref{fig:M_vx50} and \ref{fig:M_vx90}, respectively. The braking distance  depends only on $v_x$ and is the same for all models. For the largest offset of \SI{-3.7}{m}, the differences in TTC between DM and KM are \SI{200}{ms}, \SI{250}{ms}, and \SI{270}{ms} for speeds of \SI{50}{km/h}, \SI{70}{km/h}, and \SI{90}{km/h}, respectively. For the same offset, the differences in TTC between DM and DM3 are \SI{41.2}{ms}, \SI{24.1}{ms} and \SI{16.9}{ms} for speeds of \SI{50}{km/h}, \SI{70}{km/h}, and \SI{90}{km/h}, respectively.

Fig.~\ref{fig:M_psi} shows the critical zones for two different initial values of $\psi$, where
positive $\psi$ corresponds to the initial condition where the vehicle has already started steering to the left, and  negative $\psi$ when the vehicle could be coming from the adjacent lane from a previous lane change manoeuvre (for example).  It can be seen that the steering critical zone for the PMM does not change as $\psi$ changes, since the PMM does not incorporate knowledge about the yaw angle (see Section~\ref{sec:PMM}). Note that critical zones are shown only for trajectories that remain within the lane during the entire steering manoeuvre. Therefore, for DM, DM3, and SSCM, the steering critical zone starts at an offset of \SI{-3.4}{m} for $\psi(0)=\SI{-2}{\degree}$, while   it starts at a \SI{-3.5}{m} offset for $\psi(0)=\SI{2}{\degree}$.
The PMM critical zone in Fig.~\ref{fig:M_vx70_psi-5} shows that when the lateral offset is \SI{-3.7}{m} and $\psi(0)=\SI{-2}{\degree}$, the driver can still steer away with a distance of about \SI{23}{m} from the lead vehicle. In contrast, for the rest of the models, at this distance the collision can only be avoided by braking. If SSCM is applied, then the driver needs to  steer at \SI{26.1}{m} (at the latest) to avoid a crash, while if DM is applied then the distance  is \SI{27.3}{m}. This difference in distance (about \SI{1.2}{m}) is consistent for the rest of the offsets for $\psi(0)=\SI{-2}{\degree}$; see Fig.~\ref{fig:M_vx70_psi-5}. 
For $\psi(0)=\SI{-2}{\degree}$, the TTC difference between DM and KM is \SI{280}{ms} for  the largest offset, while between DM and DM3 it is \SI{27.5}{ms}.
The critical zones for DM and SSCM are similar for $\psi(0)=\SI{2}{\degree}$, while the largest difference in the distance at which steering needs to be initiated to avoid a crash  is \SI{0.8}{m}.
For $\psi(0)=\SI{2}{\degree}$, the TTC difference between DM and KM is \SI{200}{ms} for the largest offset, while between DM and DM3 it is \SI{18.8}{ms}. 

The models' critical zones for different initial lateral speeds of the ego vehicle are shown in Fig.~\ref{fig:M_vy}. The difference in the steering critical zone boundary between DM and KM is about \SI{4}{m} with respect to the initial distance and \SI{294.6}{ms} with respect to TTC for $v_\tx{s}(0)=\SI{-0.5}{m/s}$; 
see Fig.~\ref{fig:M_vx70_vy-5}. For the same conditions, the difference in TTC between the DM and DM3 is \SI{24.4}{ms}. 
For $v_\tx{s}(0)=\SI{0.5}{m/s}$, KM and PMM underestimate the required steering distance, see Fig.~\ref{fig:M_vx70_vy5}. The critical zones for SSCM and KM remain the same as in Fig.~\ref{fig:M_vx70} as $v_\tx{s}$ changes, since these two models do not include $v_\tx{s}$ in their state vectors.
The TTC difference between  models is largest between DM and KM (\SI{202.2}{ms}), while it is smallest between DM and DM3 (\SI{23.5}{ms}), for $v_\tx{s}(0)=\SI{0.5}{m/s}$ and the largest initial lateral offset.

Similarly, SSCM, KM, and PMM do not take $\dot{\psi}$ as a state, and consequently their critical zones remain the same for any $\dot{\psi}$, as shown in Fig.~\ref{fig:M_dotPsi}. As a result, the critical zone boundaries are underestimated in comparison to the DM and DM3 critical zone boundaries
, with larger differences for $\dot{\psi}(0)=\SI{-5}{\degree/s}$.
With the application of KM, TTC is underestimated  in comparison to DM by \SI{191.7}{ms} and \SI{300}{ms} for $\dot{\psi}(0)=\SI{5}{\degree/s}$ and $\dot{\psi}(0)=\SI{-5}{\degree/s}$, respectively.

Fig.~\ref{fig:M_delta} shows the critical zones for different $\delta$. The critical zone of the KM is overestimated for $\delta(0)=\SI{-2}{\degree}$ and underestimated for $\delta(0)=\SI{2}{\degree}$ with respect to the critical zone for the DM. The opposite is observed for the critical zone of the PMM, which does not change for the different $\delta$, underestimates for $\delta(0)=\SI{-2}{\degree}$, and overestimates for $\delta(0)=\SI{2}{\degree}$. The largest difference in the distance at which steering needs to be initiated to avoid a crash is observed between DM and PMM for $\delta(0)=\SI{-2}{\degree}$, which is about \SI{19}{m} and \SI{31}{m} for offsets of \SI{-2.5}{m} and \SI{0}{m}, respectively. These large differences  are because the PMM does not include $\delta$ in its state vector. The TTC difference between DM and PMM is \SI{1.35}{s}, while between DM and DM3 it is \SI{26.1}{ms} for an offset of \SI{-2.5}{m} and $\delta(0)=\SI{-2}{\degree}$.
The TTC difference between DM and KM is \SI{285.5}{ms}, while between DM and DM3 it is \SI{25.2}{ms} for an offset of \SI{-3.7}{m} and $\delta(0)=\SI{2}{\degree}$.

The computation times for different vehicle models and algorithms are shown in Table~\ref{TT:ComputationTime}.
\begin{table}[!tb]
\caption{Computation time for obtaining $t_\tx{s}$, $\mathbf x_\tx{s}$ and $x_\tx{s}$ for different models. Halley's method is used for Algorithm~\ref{alg:fy} and \ref{alg:fy_simplified}. The unit is \si{\micro s}.}
\centering
{\footnotesize
\begin{tabular}{+l^l^l^l^l}
\hline
 & DM & SSCM & KM & PM  \\
 \hline
\multicolumn{5}{l}{Algorithm~\ref{alg:fy}} \\
$t_\tx{s}$ & 686 & 429 & 417 & 79 \\
$\mathbf x_\tx{s}$ & 48 & 39 & 33 & 12 \\
$x_\tx{s}$  & 3495 & 12	& 1	& 1 \\
\rowstyle{\bfseries}
Total & 4229 &	480 &	451 &	92\\
\hline
\multicolumn{5}{l}{Algorithm~\ref{alg:fy_simplified}} \\
$t_\tx{s}$ & 686 & 429 & 417 & 79 \\
$\mathbf x_\tx{s}$ & 48 & 39 & 33 & 12 \\
$x_\tx{s}$ & 1 & 1 & 1 & 1\\
\rowstyle{\bfseries}
Total & 735 & 	469	&  451 & 92 \\
\hline
\multicolumn{5}{l}{Algorithm~\ref{alg:fy2}} \\
$t_\tx{s}$ & 1 & 1 & 1 & 1\\
$\mathbf x_\tx{s}$ & 48 & 39 & 33 & 12\\
\rowstyle{\bfseries}
Total & 49 & 40 &	34	& 13 \\
\hline
\end{tabular}}
\label{TT:ComputationTime}
\end{table}
The computation time is separated into the times needed for calculating the steering time $t_\tx{s}$, the states $\mathbf x_\tx{s}$ and the longitudinal distance $x_\tx{s}$. As can be seen, the numerical integration for computing $x_\tx{s}$ for DM using Algorithm~\ref{alg:fy} is the slowest. Here, we implement an approximate numerical integration using the trapezoidal rule with a sampling interval of \SI{0.01}{s}. For the other models, SSCM and KM, the corresponding computation times are around 290 and 3400 times shorter, respectively, because they have closed-form  analytical solutions. However, if Algorithm~\ref{alg:fy_simplified} is used, then the computation time for DM is substantially decreased. 
When  Algorithm~\ref{alg:fy2} is used for all vehicle models, the calculation of the steering time is the same for all models, since the same approximation \eqref{eq:t_s_aprox} is used. The only difference in  computation time among the models is the calculation of the states, which is always faster for the models with fewer states. For Algorithm~\ref{alg:fy2} we do not need to compute $x_\tx{s}(t_\tx{s})$, since (as mentioned earlier) this algorithm will need to be iteratively executed for different $x_\tx{s}(t_\tx{s})$. In principle, Algorithm~\ref{alg:fy2} may also be used offline in two ways. First, it is possible to start with an initial guess for $x_\tx{s}(0)$, which, for a given longitudinal speed, directly provides $x_\tx{s}(t_\tx{s})$\textemdash{}or equivalently, TTC. The corresponding lateral displacement may then provide a direction for testing different values of $x_\tx{s}(0)$ until the lateral displacement of the ego vehicle front right corner just avoids the collision at $x_\tx{s}(t_\tx{s})$. The computation time will strongly depend on the initial guess. 
To reduce this dependence, another option is to start with $x_\tx{s}(0)=0$ and then gradually increase it until the desired lateral displacement at $x_\tx{s}(t_\tx{s})$ is reached. 
For example, if the sampling interval for increasing $x_\tx{s}(0)$ is \SI{0.2}{m}, which corresponds to  a TTC interval of about \SI{0.01}{s}, then Algorithm~\ref{alg:fy2} would need to be evaluated 200 times if the sought steering time was $t_\tx{s}=\SI{2}{s}$. With a computation time of \SI{49}{\micro s} for each evaluation, Algorithm~\ref{alg:fy2} will need \SI{9800}{\micro s} to complete. This total computation time is greater than the computation time of \SI{4229}{\micro s} and \SI{735}{\micro s} for Algorithms~\ref{alg:fy} and \ref{alg:fy_simplified}, respectively, for DM. Moreover, Algorithm~\ref{alg:fy2} will give approximately the same critical zones as Algorithm~\ref{alg:fy_simplified}. Therefore, the critical zones for Algorithm~\ref{alg:fy2} are not shown. A small difference between the critical zones is to be expected, because Algorithm~\ref{alg:fy2} uses discrete samples to compute steering time. For the computational reasons mentioned above, the discrete sampling interval will likely be greater than the tolerance of \SI{e-6}{} which Algorithm~\ref{alg:fy_simplified} uses. This may slightly worsen the accuracy of Algorithm~\ref{alg:fy2} compared to Algorithm~\ref{alg:fy_simplified}. All simulations were performed on a laptop with Intel Core i7-8650 CPU at 4.2GHz with 16 GB RAM.

\section{Discussion}\label{sec:Discussion}
This paper provides a comparison of critical zones for three vehicle models described by linear steering dynamic systems (KM, SSCM, and DM) and PMM, and shows how to obtain computationally efficient solutions for the latest possible time that steering and braking need to be initiated to avoid a crash  for KM, SSCM, and DM.
The results show that, depending on the initial conditions, the critical zone is different for each of the four models. The critical zones were compared to those of the most complicated model, DM, as the ground truth is not available. Some differences between the models matter and some do not; for example, a TTC difference less than \SI{200}{ms}, despite having an effect on the intervention time,  can typically be considered of little consequence. On the contrary, a difference greater than \SI{200}{ms} may have a practical, and substantial impact on the outcome, potentially resulting in crash that another model would have avoided or resulting in unrealistically high impact speeds \cite{Brannstrom2010}. When it comes to differences in the longitudinal distance, anything greater than \SI{1}{m} can be considered to have the potential to have a practical impact on the assessment \cite{Kianfar2013}.  
The smallest differences in terms of longitudinal distance in the critical zone, less than \SI{0.38}{m}, were observed between the DM and DM3 for all tested conditions.
The smallest differences in terms of TTC, ranging between \SI{16.9}{ms} and \SI{41.2}{ms}, were observed for DM and DM3. The biggest differences for the critical zone were usually in the comparisons between KM and DM, with the TTC difference ranging from \SI{191.7}{ms} to \SI{300}{ms}. For all tested conditions, the largest difference in TTC was \SI{1.35}{s}, observed between DM and PMM when $\delta(0)=\SI{-2}{\degree}$. This large disparity is due to the fact that PMM does not change for different $\delta$, since the model does not incorporate knowledge about steering angle. However, having different $\delta$ (e.g.,  in  a previous manoeuvre the  vehicle steered to the right before  overtaking) as an initial condition would be relevant for actual assessments.

Furthermore, the times required to calculate the ADAS intervention time, for the different vehicle models and algorithms for collision avoidance by steering, were also compared. These results allow the differences to be taken into account when designing and assessing ADAS with different vehicle models. This, in turn, can facilitate quantifying the expected safety benefit of ADAS on real-world data, depending on the vehicle  model.

Regarding the different steering algorithms, the advantage of using Algorithms~\ref{alg:fy} and \ref{alg:fy_simplified} is that the critical zones can be calculated directly by backward reachability. This requires only one evaluation when the trajectory of the leading vehicle is known, for example, for performing an offline calculation of the critical zone. On the contrary, Algorithm~\ref{alg:fy2} needs to be evaluated forward for multiple longitudinal distances to find the shortest distance that avoids the collision. This algorithm might be more suitable for real-time usage on board the vehicle, as a part of a modern ADAS or higher level of automation, for example. 

In this study, a constraint (i.e., the front right vehicle corner of the ego vehicle is required to be outside the safety margin of the leading vehicle in \eqref{eq:lat-safety-margin}) is imposed only at the final time, in order to avoid a collision with the most protruding point of the leading vehicle, as illustrated in Fig.~\ref{fig:collision_avoidance}. In the future, the method can be extended by enforcing constraints at multiple points on the vehicle contour and at multiple time instances. For example, if the leading vehicle communicates its future intentions, or a better prediction can be made of the future motion of the leading vehicle. 

In this study, we demonstrated the algorithms with the assumption that the leading vehicle has a constant longitudinal speed and remains in the same lateral position. 
The proposed algorithms can easily be extended to scenarios with non-zero lateral speed of the leading vehicle (this, e.g., covers also the case where the leading vehicle motion is perpendicular to that of the ego vehicle). Since the studied systems remain linear, the computation complexity will be similar. 

The algorithms can directly deal with disturbances in, for example, positions and speeds of the leading vehicle. For offline analysis, the leading vehicle position and longitudinal speed are known from the measured data, so the algorithms will not change. For online usage, the disturbances can be taken into account by computing intervention time for the worst-case scenarios. For example, if the leading vehicle longitudinal distance and speed are uncertain but bounded, then the worst case occurs at their lower bounds. Alternatively, as the lower bounds, a common choice is taking 2-3 times the standard deviation from their mean values.

In this paper, we did not include constraints on the road or a set of final state values that ensure that the ego vehicle is positioned on the road or  parallel to the leading vehicle as in an S-manoeuvre \cite{He:2019, karlsson19}. However, the final yaw angle of the ego vehicle for all simulated conditions was between \SI{15}{\degree} and \SI{17}{\degree}, which indicates that the ego vehicle will be able to execute a mirroring manoeuvre  staying in the adjacent lane after the collision has been avoided. At these angles, the ego vehicle is laterally close to the corner of the lead vehicle, providing substantial space to recover given that there is an adjacent lane without other traffic. In the future, constraints may be included to ensure that the entire ego vehicle area stays within the road boundaries without intersecting the areas of the leading vehicle and other surrounding road users (e.g., oncoming traffic).

Furthermore, we explored the latest point in time when a driver would need to initiate braking or steering to avoid a crash, given driver comfort boundaries. The same algorithms can also be applied to compute critical zones beyond driver comfort boundaries but still within vehicle limits.

Moreover, the critical zones of the models were calculated and compared for a distinct set of initial vehicle conditions. A larger dataset that can provide a distribution of initial vehicle conditions from, for example,  naturalistic driving data, could be investigated in future studies.

The framework proposed in this study not only allows the easy exchange of vehicle models, it also allows the benchmarking of vehicle models which are described by linear steering dynamics. 
Regarding the benchmark of the  four vehicle models for calculating the  intervention time, perhaps DM needs  to be compared with more advanced and  accurate models in future studies. We demonstrated the algorithms only on straight roads, since curves were not crucial for short distances as we have in this paper. In the future, more realistic scenarios (surrounding vehicles, curved roads, intersections), and different driver models, as well as the influence of low friction, and of combined steering and braking, could be investigated.

\section{Conclusion}\label{sec:Conclusion}
In this paper, three types of single-track vehicle models and a PMM are compared in terms of critical zones describing driver comfort boundaries for avoiding collisions with a leading vehicle by braking or steering, for a specific set of vehicle conditions. 

We provide three algorithms for avoiding collision by steering. Algorithm~\ref{alg:fy} shows how to obtain a computationally efficient solution for the steering intervention time by employing Newton-Raphson and Halley’s methods. These computationally efficient methods for calculating ADAS intervention time keep the computations tractable and enable rapid safety benefit assessment. Furthermore, we provide closed-form analytical solutions for KM and SSCM.

The results show that the critical zone is different for the four models. The largest difference is observed between DM and PMM. For the given overtaking scenario and  the certain set of vehicle conditions which exemplify overtaking of a cyclist, 
the difference in TTC  between DM and PMM is about \SI{1.3}{s}, which is substantial.
If the intervention time is computed by PMM, the ADAS may underestimate the TTC at which it should start to intervene for a specific initial condition, which would result in a too late intervention.

Taking into account the small differences in the critical zones obtained  using DM with Algorithms~\ref{alg:fy} and \ref{alg:fy_simplified} and the short computation time of Algorithm~\ref{alg:fy_simplified}, it would be reasonable to use Algorithm~\ref{alg:fy_simplified} for rapid offline simulations. On the contrary, for online real-time usage in ADAS, Algorithm~\ref{alg:fy2} with DM provides further computational advantages over Algorithm~\ref{alg:fy_simplified}. 

Furthermore, the framework proposed in this paper allows for a sensitivity analysis of the critical zone with higher thresholds within the vehicle boundaries. 
Future work could extend these findings by investigating more advanced models and using  naturalistic data for validation, and may include more constraints, other surrounding vehicles, and combined steering and braking.

\appendices

\section{Lateral bicycle dynamics} \label{sec:different-models}
Different abstractions of lateral bicycle dynamics, expressed as a parameter varying state space model, 
\begin{align*}
	\dot{\mathbf x}_\tx{s} &= A_\tx{s}(v_x)\mathbf x_\tx{s} + B_\tx{s}\omega\\
	\mathbf y_\tx{s} &= C_\tx{s}(v_x)\mathbf x_\tx{s} + D_\tx{s}(v_x)\omega
\end{align*}
are summarised in Table~\ref{TT:ModelMatrices}. 

\section{Lemmas and proofs} \label{sec:lemmas}
\begin{Lemma}\label{lemma:jordan}
The solution of the ordinary linear differential equation $\dot{\mathbf x} = A\mathbf{x}$ can be written as $\mathbf x(t)=Pe^{Jt}P^{-1}\mathbf x(0)$, where $J=P^{-1}AP$ is the normal Jordan form of $A$.
\end{Lemma}
\begin{proof}
First consider the typical solution of the ordinary linear differential equation, $\mathbf x(t)=e^{At}\mathbf x(0)$. From $J=P^{-1}AP$, it follows $A=PJP^{-1}$. Now $e^{At}$ can be expanded as
\begin{align*}
\begin{split}
     e^{At}&=I+At+\frac{1}{2}A^2t^2 + \dots =\\ &PP^{-1}+PJP^{-1}t+\frac{1}{2}PJ^2P^{-1}t^2+\dots= \\
     &P(I+Jt+\frac{1}{2}J^2t^2+\dots)P^{-1} = Pe^{Jt}P^{-1}.
\end{split}
\end{align*}
Now we substitute $e^{At}$ in $\mathbf x(t)=e^{At}\mathbf x(0)$, to complete the proof, i.e., $\mathbf x(t)=Pe^{Jt}P^{-1}\mathbf x(0)$.
\end{proof}

\begin{table*}[htb]
\caption{Different model abstractions of lateral bicycle dynamics}
\centering
{\footnotesize
\begin{tabular}{l}
All vehicle models are given as $\dot{\mathbf x}_\tx{s}(t)= A_\tx{s}(\mathbf p, v_x)\mathbf x_\tx{s}(t) + B_\tx{s} u_\tx{s}$, $\mathbf y_\tx{s}(t) = C_\tx{s}(\mathbf p, v_x)\mathbf x_\tx{s}(t) + D_\tx{s}(\mathbf p, v_x) u_\tx{s}$, where $\mathbf x_\tx{s}$ are states, $u_\tx{s}$ is a control input,\\
$\mathbf y_\tx{s}$ are outputs and $\mathbf p$ is a vector of parameters. The state evolution in time is computed as $\mathbf x_\tx{s}(t)=A_{\tx{s}t}(\mathbf p, v_x,t)\mathbf x_\tx{s}(0) + B_{\tx{s}t}(\mathbf p, v_x,t)u_\tx{s}$.\\
\hline
\textbf{Dynamic model}: $\mathbf{x}_\tx{s}=\begin{bmatrix} y &  \psi &  v_\tx{s} &  \dot \psi &  \delta  \end{bmatrix}^\top$, $u_\tx{s}=\omega$, $\mathbf y_\tx{s} = \begin{bmatrix} y_\tx{FR} + \frac{W}{2} &  a_\tx{s} &  j_\tx{s} \end{bmatrix}^\top$, $\delta_\tx{ss}(a_\tx{smax},v_x) = \frac{a_\tx{smax}}{l}\left(\left(\frac{l}{v_x}\right)^2 + \frac{m}{2}\left(\frac{l_\tx{r}}{c_\tx{f}} - \frac{l_\tx{f}}{c_\tx{r}} \right)\right)$,\\
$\omega_\tx{ss}(j_\tx{smax},v_x)= \frac{j_\tx{smax}}{l}\left(\left(\frac{l}{v_x}\right)^2 + \frac{m}{2}\left(\frac{l_\tx{r}}{c_\tx{f}} - \frac{l_\tx{f}}{c_\tx{r}} \right)\right)$, $\mathbf p=2\begin{bmatrix} \frac{c_\tx{f}+c_\tx{r}}{m} & \frac{l_\tx{r}c_\tx{r}-l_\tx{f}c_\tx{f}}{m} & \frac{c_\tx{f}}{m} & \frac{l_\tx{r}c_\tx{r}-l_\tx{f}c_\tx{f}}{I_z} & \frac{l_\tx{f}^2 c_\tx{f}+l_\tx{r}^2 c_\tx{r}}{I_z} & \frac{l_\tx{f} c_\tx{f}}{I_z} & \frac{L_\tx{f}}{2} \end{bmatrix}^\top$,\\
$A_\tx{s}=\begin{bmatrix}0 & v_x & 1 & 0 & 0\\
                      0 & 0 & 0 & 1 & 0 \\
                     0 & 0 & -\frac{p_1}{v_x} & \frac{p_2}{v_x} - v_x & p_3\\
                     0 & 0 & \frac{p_4}{v_x} & -\frac{p_5}{v_x} & p_6 \\
                     0 & 0 & 0 & 0 & 0
    \end{bmatrix}$,
 $B_\tx{s}=\begin{bmatrix}0 \\ 0 \\ 0 \\ 0\\ 1 \end{bmatrix}$,
 $C_\tx{s}=\begin{bmatrix}
     1 & p_7 & 0 &                 0 &                       0\\ 
     0 & 0 & -\frac{p_1}{v_x} &         \frac{p_2}{v_x} &        p_3\\ 
     0 & 0 & \frac{p_1^2 + p_2 p_4}{v_x^2} & p_1 - p_2\frac{p_1 + p_5}{v_x^2} & \frac{p_2 p_6 - p_1p_3}{v_x}
    \end{bmatrix}$,
 $D_\tx{s}=\begin{bmatrix}0 \\ 0 \\ p_3 \end{bmatrix}$.\\
 \hline
 \textbf{Steady-state cornering model}: $\mathbf{x}_\tx{s}=\begin{bmatrix} y & \psi & \delta \end{bmatrix}^\top$, $u_\tx{s}=\omega$, $\mathbf y_\tx{s} = \begin{bmatrix} y_\tx{FR} + \frac{W}{2} &  a_\tx{s} &  j_\tx{s} \end{bmatrix}^\top$,\\
$\mathbf p=\begin{bmatrix} l_\tx{r} & l & \frac{m l_\tx{f}}{2c_\tx{r}l} & \frac{m}{2 l}\left( \frac{l_\tx{r}}{c_\tx{f}} - \frac{l_\tx{f}}{c_\tx{r}} \right) & L_\tx{f} \end{bmatrix}^\top$, $\delta_\tx{ss}(a_\tx{smax}, v_x)=a_\tx{smax}(\frac{p_2}{v_x^2}+p_4)$, $\omega_\tx{ss}(j_\tx{smax}, v_x)=j_\tx{smax}(\frac{p_2}{v_x^2}+p_4)$, $v_\tx{s} = \frac{p_1-p_3 v_x^2}{p_2+p_4 v_x^2} v_x \delta$.\\
 $A_\tx{s} =\begin{bmatrix}
    0 & v_x & \frac{p_1 - p_3 v_x^2}{p_2+p_4v_x^2}v_x\\
    0 & 0 & \frac{v_x}{p_2+p_4v_x^2} \\
     0 & 0 & 0 
    \end{bmatrix}$,
    $B_\tx{s}=\begin{bmatrix}0 \\ 0 \\ 1 \end{bmatrix}$,
    $C_\tx{s} =\begin{bmatrix}
    1 & p_5 & 0 \\
    0 & 0 & \frac{v_x^2}{p_2+p_4v_x^2} \\
    0 & 0 & 0
    \end{bmatrix}$,
    $D_\tx{s} =\begin{bmatrix}0 \\ 0 \\ \frac{v_x^2}{p_2+p_4v_x^2} \end{bmatrix}$.\\
    $A_\tx{st} =\begin{bmatrix}
    1 & v_x t & \frac{v_x^2 t^2 + 2 v_x t(p_1 - p_3 v_x^2)}{2(p_2+p_4v_x^2)}\\
    0 & 1 & \frac{v_x t}{p_2+p_4v_x^2} \\
    0 & 0 & 1 
    \end{bmatrix}$,
    $B_\tx{st}=\begin{bmatrix}\frac{v_x^2 t^3 + 3v_x t^2 (p_1-p_3v_x^2)}{6(p_2 +p_4 v_x^2)} \\ \frac{v_x t^2}{2(p_2 +p_4 v_x^2)} \\ t \end{bmatrix}$.\\
    $x_\tx{s}(t_\tx{s}) = x_\tx{s}(0) + v_x t_\tx{s} - \frac{t_\tx{s} v_x (2 \delta(0) + \omega_\tx{max} t_\tx{s}) (p_1-p_3 v_x^2) (\omega_\tx{max} t_\tx{s}^2 v_x + 2 \delta(0) t_\tx{s} v_x + 4 p_4 \psi(0) v_x^2 + 4 p_2 \psi(0)) }{8(p_4 v_x^2 + p_2)^2}$ for $t_\tx{sa} \geq t_\tx{s}$, otherwise\\
    $x_\tx{s}(t_\tx{s}) = x_\tx{s}(0) + v_x t_\tx{s} - \frac{t_\tx{sa} v_x (2 \delta(0) + \omega_\tx{max} t_\tx{sa}) (p_1-p_3 v_x^2) (\omega_\tx{max} t_\tx{sa}^2 v_x + 2 \delta(0) t_\tx{sa} v_x + 4 p_4 \psi(0) v_x^2 + 4 p_2 \psi(0)) }{8(p_4 v_x^2 + p_2)^2}$\\
    $+ \frac{\delta_\tx{max} v_x (p_1-p_3 v_x^2) (t_\tx{sa}-t_\tx{s}) (2 p_2 \psi(0) + 2p_4 \psi(0) v_x^2 + \omega_\tx{max} t_\tx{sa}^2 v_x + 2 \delta(0) t_\tx{sa} v_x - \delta_\tx{max} t_\tx{sa} v_x +\delta_\tx{max} t_\tx{s} v_x) }{2(p_4 v_x^2 + p_2)^2}$\\
\hline 
 \textbf{Kinematic model}: $\mathbf{x}_\tx{s}=\begin{bmatrix} y & \psi & \delta \end{bmatrix}^\top$, $u_\tx{s}=\omega$, $\mathbf y_\tx{s} = \begin{bmatrix} y_\tx{FR} + \frac{W}{2} &  a_\tx{s} &  j_\tx{s} \end{bmatrix}^\top$, $\mathbf p=\begin{bmatrix} \frac{l_\tx{r}}{l} & \frac{1}{l} & L_\tx{f} \end{bmatrix}^\top$, $\delta_\tx{ss}(a_\tx{smax}, v_x)=\frac{a_\tx{smax}}{p_2v_x^2}$,  $\omega_\tx{ss}(j_\tx{smax}, v_x)=\frac{j_\tx{smax}}{p_2v_x^2}$,\\
 $v_\tx{s} = \frac{l_\tx{r}}{l}v_x \delta$.\\
$A_\tx{s}=\begin{bmatrix}
    0 & v_x & p_1 v_x\\
    0 & 0 & p_2 v_x \\
     0 & 0 & 0 
    \end{bmatrix}$,
    $B_\tx{s}=\begin{bmatrix}0 \\ 0 \\ 1 \end{bmatrix}$,
    $C_\tx{s}=\begin{bmatrix}
    1 & p_3 & 0\\
    0 & 0 & p_2 v_x^2 \\
    0 & 0 & 0 
    \end{bmatrix}$,
    $D_\tx{s}=\begin{bmatrix}0 \\ p_1 v_x \\ p_2 v_x^2 \end{bmatrix}$,
    $A_\tx{st}=\begin{bmatrix}
    1 & v_x t & v_x (p_1 t + \frac{p_2 v_x t^2}{2}) \\
    0 & 1 & p_2 v_x t \\
    0 & 0 & 1 
    \end{bmatrix}$,
    $B_\tx{st}=\begin{bmatrix} \frac{1}{6}p_2 v_x^2 t^3+\frac{1}{2} p_1 v_x t^2 \\ \frac{1}{2}p_2 v_x t^2 \\ t \end{bmatrix}$.\\
    $x_\tx{s}(t_\tx{s}) = x_\tx{s}(0) + v_x t_\tx{s} - \frac{l_r t_\tx{s} v_x (2\delta(0) +\omega_\tx{max} t_\tx{s})(\omega_\tx{max} p_2 v_x t_\tx{s}^2 + 2\delta(0) p_2 v_x t_\tx{s} + 4 \psi(0))}{8l}$ for $t_\tx{sa} \geq t_\tx{s}$, otherwise\\
    $x_\tx{s}(t_\tx{s}) = x_\tx{s}(0) + v_x t_\tx{s} - \frac{l_r t_\tx{sa} v_x (2\delta(0) +\omega_\tx{max} t_\tx{sa})(\omega_\tx{max} p_2 v_x t_\tx{sa}^2 + 2\delta(0) p_2 v_x t_\tx{sa} + 4 \psi(0))}{8l}$ \\
    $+ \frac{\delta_\tx{max} l_r v_x (t_\tx{sa}-t_\tx{s}) (2\psi(0) + 2\delta(0) p_2 t_\tx{sa} v_x - \delta_\tx{max} p_2 t_\tx{sa} v_x +\delta_\tx{max} p_2 t_\tx{s} v_x + \omega_\tx{max} p_2 t_\tx{sa}^2 v_x) }{2l} $\\
\hline 
 \textbf{Point mass model}: $\mathbf{x}_\tx{s}=\begin{bmatrix} y & v_\tx{s} & a_\tx{s} \end{bmatrix}^\top$, $u_\tx{s}=j_\tx{s}$, $\mathbf y_\tx{s} = \begin{bmatrix} y_\tx{FR} + \frac{W}{2} &  a_\tx{s} &  j_\tx{s} \end{bmatrix}^\top$,\\
$A_\tx{s}=\begin{bmatrix}
    0 & 1 & 0 \\
    0 & 0 & 1  \\
    0 & 0 & 0 
    \end{bmatrix}$,
    $B_\tx{s}=\begin{bmatrix}0 \\ 0 \\ 1 \end{bmatrix}$, 
    $C_\tx{s}=\begin{bmatrix}
    1 & 0 & 0 \\
    0 & 0 & 1  \\
    0 & 0 & 0 
    \end{bmatrix}$, 
    $D_\tx{s}=\begin{bmatrix}0 \\0 \\ 1 \end{bmatrix}$,
    $A_\tx{st}=\begin{bmatrix}
    1 & t & \frac{t^2}{2} \\
    0 & 1 & t  \\
    0 & 0 & 1 
    \end{bmatrix}$,
    $B_\tx{st}=\begin{bmatrix}\frac{t^3}{6} \\ \frac{t^2}{2} \\ t \end{bmatrix}$.\\
    $x_\tx{s}(t_\tx{s}) = x_\tx{s}(0) + v_x t_\tx{s}$\\
\hline
\end{tabular}}
\label{TT:ModelMatrices}
\end{table*}

\bibliographystyle{IEEEtran}
\bibliography{IEEEabrv,biblio}
\vfill
\end{document}